\documentclass[runningheads]{llncs}

\usepackage{eccv}

\usepackage{eccvabbrv}

\usepackage{url}

\usepackage{graphicx}
\usepackage{amsmath}
\usepackage{amssymb}
\usepackage{booktabs}
\usepackage{multirow}
\usepackage{xcolor}
\usepackage{float}

\usepackage{comment}
\usepackage{amsmath,amssymb} 
\usepackage{color}

\usepackage{bbold}
\usepackage[accsupp]{axessibility}  

\usepackage{hyperref}

\usepackage{orcidlink}

\begin{document}

\title{On the Viability of Monocular Depth Pre-training for Semantic Segmentation} 

\titlerunning{On Viability Monocular Depth Pre-train. Semantic Segmentation}

\author{Dong Lao \inst{1}\orcidlink{0000-0001-9308-7085} \and
Fengyu Yang \inst{2}\orcidlink{0009-0001-1094-8204} \and
Daniel Wang\inst{2}\orcidlink{0009-0009-5737-4699} \and
Hyoungseob Park\inst{2}\orcidlink{0000-0003-0787-2082} \and
Samuel Lu \inst{1}\orcidlink{0009-0003-4334-1374} \and
Alex Wong\inst{2}\orcidlink{0000-0002-3157-6016} \and
Stefano Soatto\inst{1}\orcidlink{0000-0003-2902-6362}}

\authorrunning{D Lao et al.}

\institute{UCLA Vision Lab, Los Angeles, CA 90024, USA \\
\email{\{lao, soatto\}@cs.ucla.edu, samuellu@ucla.edu}\ \and
Yale Vision Lab, New Haven, CT 06511, USA \\
\email{\{fengyu.yang, daniel.wang.dhw33, hyoungseob.park, alex.wong\}@yale.edu}}

\maketitle

\begin{abstract}
The question of whether pre-training on geometric tasks is viable for downstream transfer to semantic tasks is important for two reasons, one practical and the other scientific. If the answer is positive, we may be able to reduce pre-training costs and bias from human annotators significantly. If the answer is negative, it may shed light on the role of embodiment in the emergence of language and other cognitive functions in evolutionary history. To frame the question in a way that is testable with current means, we pre-train a model on a geometric task, and test whether that can be used to prime a notion of “object” that enables inference of semantics as soon as symbols (labels) are assigned.  We choose monocular depth prediction as the geometric task, and semantic segmentation as the downstream semantic task, and design a collection of empirical tests by exploring different forms of supervision, training pipelines, and data sources for both depth pre-training and semantic fine-tuning. We find that monocular depth \emph{is} a viable form of pre-training for semantic segmentation, validated by improvements over common baselines. Based on the findings, we propose several possible mechanisms behind the improvements, including their relation to dataset size, resolution, architecture, in/out-of-domain source data, and validate them through a wide range of ablation studies. We also find that optical flow, which at first glance may seem as good as depth prediction since it optimizes the same photometric reprojection error, is considerably less effective, as it does not explicitly aim to infer the latent structure of the scene, but rather the raw phenomenology of temporally adjacent images. Code: \url{https://github.com/donglao/DepthToSemantic}.
\end{abstract}

\section{Introduction}
We probe the following seemingly counter-intuitive hypothesis: \\
\centerline{\emph{Can pre-training on a geometric task benefit a downstream semantic task?}}\\
Geometric inference is often viewed as a low-level vision task requiring little abstraction that is needed for semantics~\cite{julesz1971foundations}. For example, depth can be acquired through minimizing reprojection error, \ie from multi-view or videos, or directly from range sensors. Both can be performed procedurally, without inductive learning, rendering depth a meaningless task for pre-training. However, induction is needed to infer one 3D scene among infinitely many compatible with the same 2D image. Therefore, if a model could solve this ill-posed problem, it would provide evidence of the viability of pre-training with little to no human intervention, which is important in specialized data domains for which little annotated data is publicly available. 

In this paper, we focus on testing monocular depth as the pre-training geometric task, and semantic segmentation as the downstream semantic task. They are purposefully chosen: Training deep neural networks for semantic segmentation requires labor-intensive pixel-level annotation, so the choice of pre-training is essential to its performance. Existing studies have shown mixed results about the relationship between the two tasks. Taskonomy \cite{zamir2018taskonomy}, a framework for measuring relationships between visual tasks, suggests that depth estimation is ``far'' from semantic segmentation, while recent work \cite{goldblum2024battle} shows that depth pre-training can beat ``closer'' tasks like image classification. Prior work \cite{jiang2018self,ramirez2019learning,hoyer2021three,saha2021learning} has also shown improvement in semantic segmentation when incorporating depth. This mixed evidence motivates us to take a closer look at the underlying mechanism of how monocular depth may benefit semantic segmentation.

Another more subtle reason for exploring this hypothesis is that pre-training is often performed on heavily human-biased datasets \cite{imagenet_cvpr09, oquab2024dinov}, where the photographer who framed the picture meant to convey a particular concept (say, a cup), and therefore took care to make sure that the manifestation of the concept (the image) prominently features the object by choice of vantage point, illumination, and (lack of) occlusion. This bias is mitigated if data is not purposefully organized into ``shots.'' Unfortunately, existing datasets are mostly composed of purposefully framed shots which could obfuscate the analysis. We note that depth can be inferred without any semantic interpretation \cite{julesz1971foundations} regardless of whether the data is captured purposefully or randomly. With monocular depth as the pre-training task, there are two ways of reducing the aforementioned human selective bias: The first involves directly pre-training within the specific domain of interest, leveraging the simplicity of data gathering; The second way is scaling up pre-training by incorporating diverse sources of data, which is made possible by recent developments \cite{lasinger2019towards,ranftl2021vision, yang2024depth} on relative depth estimation.

{\bf \noindent Methods.} We formalize the main hypothesis in Sect. \ref{sec:formalization}. Since it cannot be tested analytically without knowledge of the joint distribution of test images and labels, we propose an empirical testing protocol. We test on monocular depth models trained under multiple forms of supervision, including structure-from-motion, binocular stereo, and depth sensors. We then change the prediction head of the resulting network, either the final layer or the whole decoder, and fine-tune it for semantic segmentation (Fig.~\ref{fig:diagram}). We consider depth estimation as a \textit{viable} pre-training option if it yields comparable improvements to downstream semantic segmentation tasks as other common pre-training practices, \eg ImageNet classification. To this end, we design a series of controlled experiments to test the effect of choice of initialization (\cref{tab:results}, \cref{fig:comparison}), training with various datasets sizes (\cref{fig:trainsize}), choice of network component to be frozen and fine-tuned (\cref{fig:fixed}), effect of resolution of training images (\cref{fig:scaling}). Conclusions are drawn from both quantitative and qualitative (\cref{fig:activation}) results.

{\bf \noindent Findings.} Pre-training for depth estimation improves the performance of downstream semantic segmentation across different experimental settings. Particularly, we show that depth estimation is indeed a viable pre-training option as compared to existing methods (\cref{tab:self_supervised}). For example, compared to classification, using depth on average improves by 5.8\% mIoU and 5.2\% pixel accuracy on KITTI. As a sanity check, we test both a depth network pre-trained from scratch and one trained after ImageNet initialization, and both outperform classification-based pre-training in downstream semantic segmentation. To control the effect of our choice of architecture, we used our pre-trained encoder to initialize a standard semantic segmentation network \cite{chen2017deeplab}. We observed similar findings on Cityscapes and NYU-V2 regardless of how depth training is supervised. Inferring depth without explicit supervision typically involves minimizing the prediction error, just like optical flow. Somewhat surprisingly, not only does pre-training for depth outperform optical flow, but the latter is often worse than random initialization (Fig.~\ref{fig:flow}). One may also argue that observed improvements mainly come from the availability of in-domain pre-training data for depth. To test this conjecture, we fine-tune a depth model \cite{depthanything} trained on large-scale out-of-domain data. Improvements in semantic segmentation reveal that when trained at scale, depth models show strong transferability to unseen downstream data domains.

\section{Related Work}
\label{sec:related}

Pre-training aims to learn a representation (function) of the {\em test} data that is maximally informative (sufficient), while providing some kind of complexity advantage. In our case, we measure complexity by the validation error after fine-tuning on limited amount of labeled data, which measures the inductive value of pre-training. The recent literature comprises a large variety of ``self-supervised'' methods that are purportedly task-agnostic. In reality, the task is specified indirectly by the choice of hand-designed nuisance transformations that leave the outcome of inference unchanged. Such transformations are sampled through data augmentation while the image identity holds constant (Contrastive Learning) \cite{chen2020improved,chen2020simple,caron2021emerging,oquab2024dinov}, or reconstruction \cite{Caron_2019_ICCV}. Group transformations organize the dataset into orbits, which contrastive learning tries to collapse onto its quotient, which is a maximal invariant. Such a maximal invariant is transferable to all and only tasks {\em for which the chosen transformation is uninformative}. For group transformation, the maximal invariant can, in theory, be computed in closed form \cite{sundaramoorthi2009set}. In practice, contrastive learning are extended to non-group transformations, \eg occlusions, as seen in language \cite{brown2020language} and images \cite{chen2019unsupervised}. All self-supervised methods boil down to hand-designed and quantized subsets of planar domain diffeomorphisms (discrete rotations, translations, scaling, reflections, \etc), range homeomorphisms (contrast, colormap transformations) and occlusion masks. 

In our case, rather than hand-designing the nuisance transformations assumed to be shared among pre-training and fine-tuning tasks, we let the scene itself provide the needed supervision: images portend the same scene, either from the same timestamp (stereo) or adjacent in the temporal domain (video frames), so their variability defines the union of nuisance factors. These include domain deformations due to ego- and scene motion, range transformations due to changes in illumination, and occlusions. In addition to sharing nuisance variability, pre-training and fine-tuning tasks should ideally also share the hypothesis space. It may seem odd to choose a geometric task, where the hypothesis space is depth, to pre-train for a semantic task, where the hypothesis space is a discrete set of labels. However, due to the statistics of range images \cite{huang2000statistics} and their similarity to the statistics of natural images \cite{huang1999statistics}, this is actually quite natural: A range map is a piecewise smooth function defined on the image domain, whereas a segmentation map is a piecewise constant function where the levels are mapped to arbitrary labels. As a result, the decoder for depth estimation can be easily modified for semantic segmentation. A discussion of this choice, specifically on the representational power of deterministic predictors, in Sect. \ref{sec:discussion}. 
\cite{jiang2018self,hoyer2021three} also utilize depth for semantic segmentation. \cite{jiang2018self} proposes pre-training on relative depth prediction, and \cite{hoyer2021three,hoyer2021improving} utilizes self-supervised depth estimation on video sequences. Our experiments validate their findings. We further investigate whether features obtained purely from monocular depth improve semantic segmentation.

Monocular depth \cite{monodepth2,wong2019bilateral} may use different supervision, either through additional sensors \cite{fei2019geo,wong2021unsupervised,wu2024augundo}, or synthetic data \cite{wong2021learning,lopez2020project,yang2019dense}, but none require human annotation. Some use regularizers with sparse seeds \cite{wong2020unsupervised,liu2022monitored,wong2021adaptive}, or adopt pre-trainings \cite{ranftl2021vision}.  
We design experiments agnostic to how depth models are trained, but also make comparisons across different forms of depth supervision (Tab.~\ref{tab:supervision}).

\section{Formalization}
\label{sec:formalization}

\begin{figure*}[t]
\centering
\vspace{-0.5em}
\includegraphics[width=0.85\textwidth]{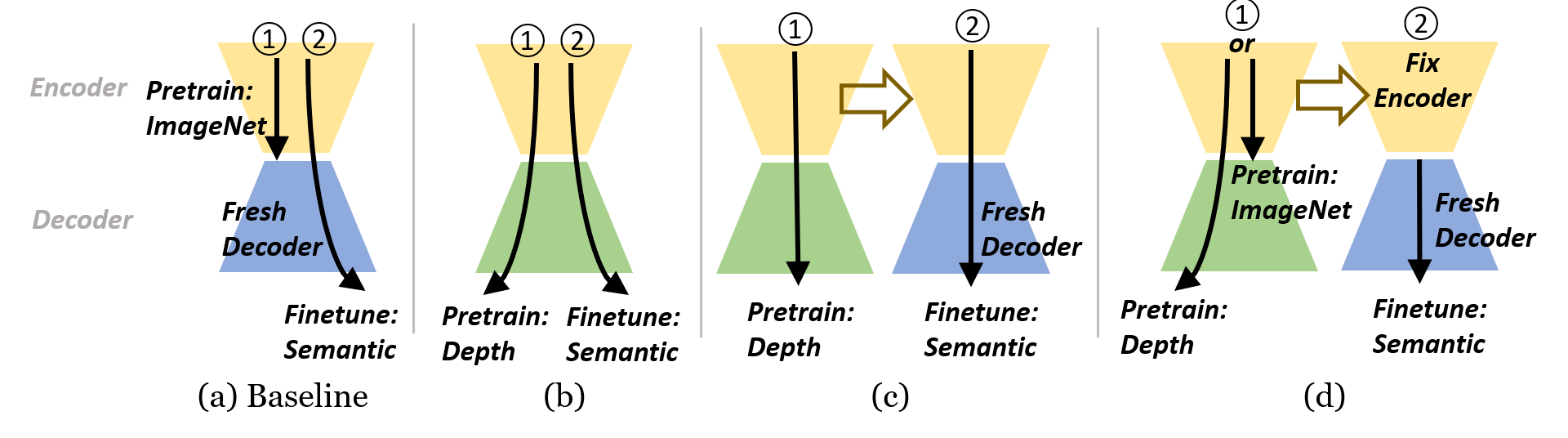}

    \caption{{\bf Diagram for different pre-training and fine-tuning setups.} (a) Common practice: pre-train the encoder, \eg on ImageNet, attach a decoder, and fine-tune the network. (b) Our best practice: pre-train the network by monocular depth, and fine-tune for semantic segmentation. (c) Cross architecture: for fair comparisons with common practice, we pre-train by depth, replace the decoder, and fine-tune. (d) To test the quality of pre-trained encoders, we fix the encoders and fine-tune decoders only.}
    \label{fig:diagram}

\end{figure*}

\def\x{{\bf{x}}}
\def\y{{\bf{y}}}
\def\z{{\bf{z}}}
\def\h{{\bf{h}}}

Let $x: D\subset {\mathbb R}^2 \rightarrow \{0, \dots, 255\}^3$ be an image, where the domain $D$ is quantized into a lattice, $z: D \rightarrow \{1, \dots, Z\}$ a depth map with $Z$ depth or disparity levels, and $y: D \rightarrow \{1, \dots, K\}$ a semantic segmentation map. In coordinates, each pixel in the lattice, $(i,j) \in \{1, \dots, N\} \times \{1, \dots, M\}$ is mapped to RGB intensities by $x(i,j)$, a depth by $z(i,j)$, and a label by $y(i,j)$. Despite the discrete nature of the data and the hypothesis space, we relax them to the continuum by considering the vectors $\x \in {\mathbb R}^{NM3}$, $\y \in {\mathbb R}^{NMK}$ and $\z \in {\mathbb R}^{NM}$. With a slight abuse of notation, $y \in \{1, \dots, K\}$ denotes a single label and $\bar y \in {\mathbb R}^K$ its embedding, often restricted to a one-hot encoding.

Now, consider a dataset ${\cal D}_z = \{\x_t^i, \z_t^i \}_{i,t = 1}^{V, T_i}$ comprised of $V$ image sequences each of length $T_i$. For the simplicity of the notations, in the case of multi-view stereo, we also consider multiple 2D image inputs as a ``sequence'' without loss of generality. In the case of supervised depth estimation, synchronized depth maps $z_t^i$'s are measured by a range sensor. Typical datasets supporting depth estimation may include just image sequences or both modalities.

Training for monocular depth estimation yields a mapping $\phi_w: \x \mapsto \z$, parametrized by weights $w$ in a neural network, via
\begin{equation}\label{eq:loss}
   w = \arg\min_{w,g_t}\sum_{i,j, n, t}
   \ell(x^n_{t+1}(i,j), \hat x^n_t(i,j)) \\
   +\lambda \sum_{i,j, n, t}
   \ell(z^n_{t}(i,j), \hat z^n_t(i,j))
\end{equation}
where $\hat x_t$ is the warping of an image $x$ from $t$ to $t+1$ based on camera pose $g_t$. Here we consider a generic formulation for different modalities of depth estimation. The first term in \cref{eq:loss} measures reprojection error across frames, and the second term measures the distance between estimated depth values and the ground-truth from the range sensor. In the case of unsupervised depth estimation from videos, \eg \cite{monodepth2}, only the reprojection loss is considered; while in the case of supervised depth estimation from single images, \eg \cite{NYUV2}, $T_i=1$ for all $i$'s, so only the second term is minimized.

The goal is to use these representations as encodings of the data to then learn a semantic segmentation map. In practice, the representations above are implemented by deep neural networks, that can be truncated at intermediate layers thus providing embedding spaces larger than the respective hypothesis spaces. We refer to the parts before and after this intermediate layer {\em encoder} and {\em decoder}, respectively. We overload the notation and refer to the encoding as $\h = \phi_w(\x)$ for both depth estimation and  other pre-training methods, presumably with weights $w'$, assuming they have the same encoder architecture. The goal of semantic segmentation is then to learn a parametrized map $\psi_{w''}: \h \mapsto \y$ using a small but fully supervised dataset  ${\cal D}_{s} = \{\x^n, \y^n\}_{n=1}^N$, by minimizing some loss function or (pseudo-)distance in the hypothesis space $d(\y, \hat \y)$, where
 \begin{equation}
     w'' = \arg\min_w \sum_{n=1}^N d(\y^n, \psi_w(\h^n))
 \end{equation}
plus customary regularizers. In the aggregate, we have a Markov chain:
$
    \x \longrightarrow \h = \phi_w(\x) \longrightarrow \y = \psi_{w''}(\h) = \psi_{w''}\circ \phi_w(\x)
$
for depth estimation, and 
$
    \x \longrightarrow \hat \h = \phi_{w'}(\x) \longrightarrow \y =  \psi_{w''}\circ \phi_{w'}(\x)
    $
for other pre-training methods. A representation obtained through a Markov chain is optimal (minimal sufficient) only if the intermediate variable $\h$ or $\hat \h$ reduces the Information Bottleneck \cite{tishby2000information} to zero. In general, there is information loss, so we {\bf formalize the key question} 
as whether the two Information Bottleneck Lagrangians satisfy the following:
\begin{equation}
   H(\y| \h) + \beta I(\h; \x) \stackrel{?}{\leq}  H(\y| \hat \h) + \beta' I(\hat \h; \x)
   \label{eq:question}
\end{equation}
where $\beta$ and $\beta'$ are hyperparameters that can be optimized as part of the training process, and $I, H$ denotes the (Shannon) Mutual Information and cross-entropy respectively. If the above is satisfied, then pre-training for depth estimation is a viable option, or even better than pre-training with another method. It would be ideal if this question could be settled analytically. Unfortunately, this is not possible, but the formalization above suggests a protocol to settle it empirically.

To test this empirically, we use the validation error on a supervised dataset $\mathcal{D}_s$ as a proxy for residual information. We conduct fine-tuning under several configurations (Fig.~\ref{fig:diagram}): with respect to $w''$ using $\mathcal{D}_s$, \ie yielding a comparison of the raw pre-trained back-bones (encoders, $w, w'$), or with respect to {\em both} $w''$ {\em and} $w$ (for depth estimation) or $w'$ (for other pre-training methods). Finally, all four resulting models can be compared with one obtained by training from scratch by optimizing a generic architecture with respect to $w''$ alone.

\begin{table}[tb]
  \caption{{\bf Semantic segmentation accuracy on KITTI.} Unsupervised depth as pre-training improves semantic segmentation accuracy under all settings. Our best practice (in {\bf\color{blue} blue}) improves common practice (in {\bf\color{purple}purple}) by 7.53\% mIoU and 4.68\% pixel accuracy. Freezing the encoder with ImageNet pre-training (in {\bf\color{red}red}) is worse than no pre-training (random initialization). DeepLabV3$^\dag$: with ResNet50 encoder. 
  }
    \label{tab:results}
  \centering
  \scalebox{1}{
\footnotesize
  \begin{tabular}{l|cc|cc|cc|cc|cc}
    &\multicolumn{6}{c|}{Fine-tune All}&\multicolumn{4}{c}{Freeze Encoder} \\ \hline
    &\multicolumn{2}{c|}{ResNet18}&\multicolumn{2}{c|}{ResNet50}&\multicolumn{2}{c|}{DeepLabV3$^\dag$}&\multicolumn{2}{c|}{ResNet18}&\multicolumn{2}{c}{ResNet50}\\    \hline
    Pre-training&mIoU&P.Acc&mIoU&P.Acc&mIoU&P.Acc&mIoU&P.Acc&mIoU&P.Acc
    \\
    \hline
    None &41.35&70.75&44.66&73.37&21.93&52.32&41.24&70.52&37.72&67.38\\
 
    ImageNet&45.15&72.39&44.65&73.06&\color{purple}\bf{43.39}&\color{purple}\bf{72.66}&\color{red}\bf{33.33}&\color{red}\bf{65.34}&\color{red}\bf{32.03}&\color{red}\bf{62.53}\\

    Depth-Rand &46.00&72.43&49.90&76.28&43.43&71.34&43.02&72.38&45.79&\bf{74.71}\\
    
    Depth &\bf{50.20}&\bf{76.39}&\color{blue}\bf{50.92}&\color{blue}\bf{77.34}&\bf{43.77}&\bf{72.68}&\bf{46.53}&\bf{74.42}&\bf{46.55}&74.48
  \end{tabular}}

\end{table}

\section{Experiments}


\subsection{Controlled Experiments with Few-shot Fine-tuning}
\label{sec:setups}
\def\figd{figures/plots}
\def\fWidD{0.24\textwidth}
\begin{figure}[t]

\centering
{\footnotesize
\begin{tabular}{c@{\hspace{0.03in}}c@{\hspace{0.03in}}c@{\hspace{0.03in}}c}
\rotatebox{90}{\quad ResNet18}&\includegraphics[width=\fWidD]{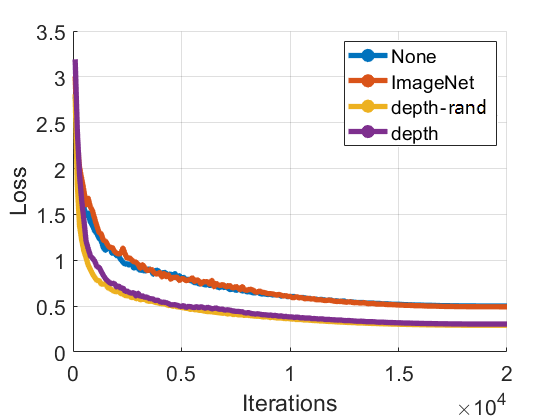} & \includegraphics[width=\fWidD]{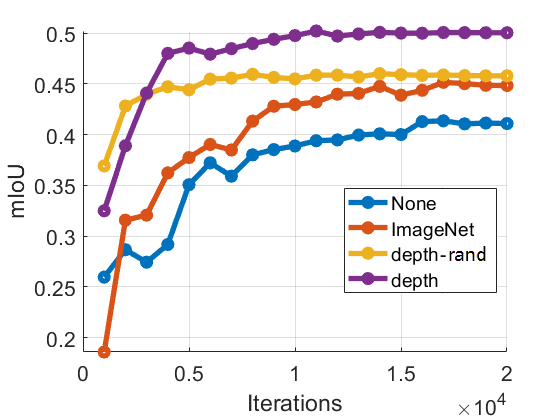} &
\includegraphics[width=\fWidD]{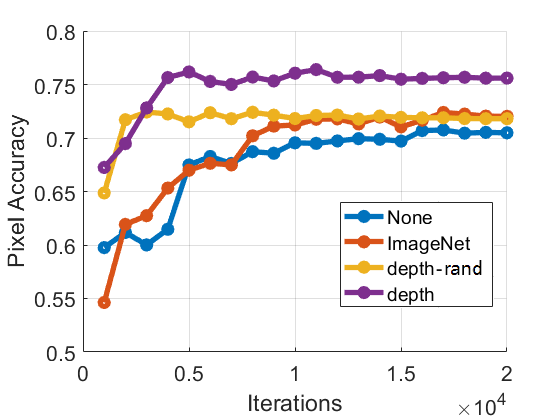} \\
\rotatebox{90}{\quad ResNet50}&\includegraphics[width=\fWidD]{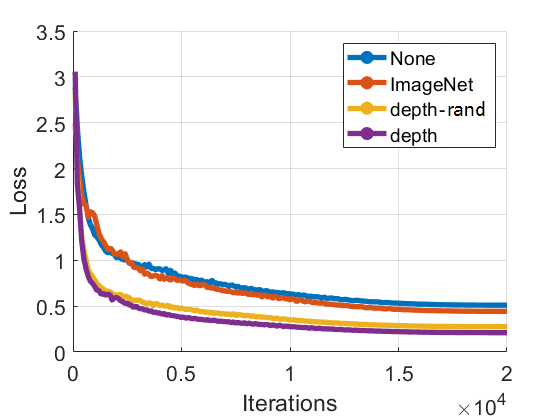} & \includegraphics[width=\fWidD]{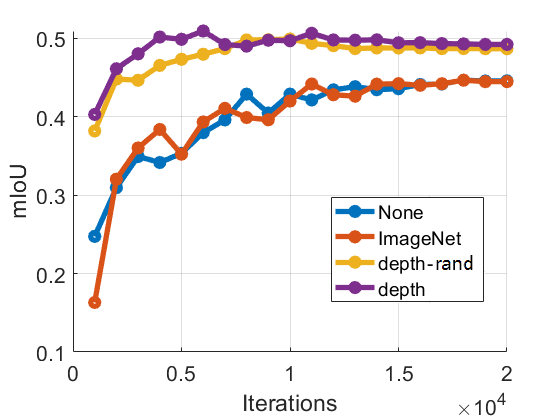} &
\includegraphics[width=\fWidD]{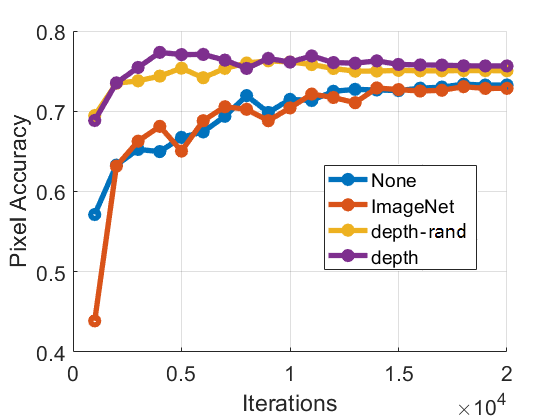}
\end{tabular}
}

    \caption{ {\bf Comparison between different network initializations.} Models initialized by depth pre-training (unsupervised) train faster and achieve higher final accuracy.}

\label{fig:comparison}
\end{figure}

We first cover an extensive collection of controlled experiments and ablations to gain insights into the main hypothesis. We specifically conduct experiments under the \emph{few-shot} setting, where only a small amount of labels are available for fine-tuning, to highlight the role of pre-training.

\begin{figure}[t]
\begin{minipage}[b]{.52\textwidth}
\centering
\def\fWidD{0.48\textwidth}
{\footnotesize
\def\figd{figures/plots}
\begin{tabular}{c@{\hspace{0.05in}}c}
\includegraphics[width=\fWidD]{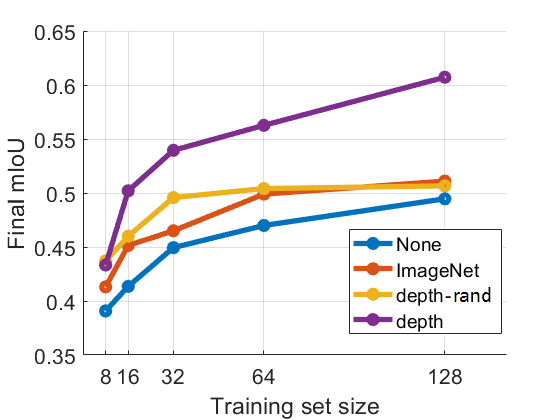}
&\includegraphics[width=\fWidD]{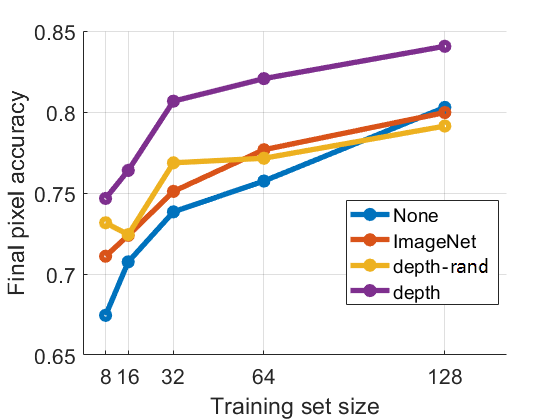} 
\end{tabular}
}

\caption{ {\bf Final accuracy vs different training set size.} Under all training set sizes, our best practice constantly outperforms ImageNet pre-trained. Encoder: ResNet 18.}

\label{fig:trainsize}
\def\figd{figures/plots}
\end{minipage}
\hfill
\begin{minipage}[b]{.45\textwidth}
\centering
    \begin{tabular}{l|c|c}
        Depth Sup. & mIoU & P.Acc \\\hline
        Video & 46.00 & 72.43 \\\hline
        Stereo & 49.11 & 74.58 \\\hline
        Lidar & \bf{52.78} & \bf{77.17}
    \end{tabular}

\captionof{table}{ {\bf Forms of depth supervision matter.} Direct supervision with Lidar works the best, followed by stereo (with known camera pose), and monocular video (camera pose unknown).}\label{tab:supervision}
\end{minipage}
\end{figure}

{\bf KITTI~\cite{KITTI}} contains 93000 video frames for depth training with 200 densely annotated images for semantic segmentation. Segmentation results are evaluated by mean IoU (mIoU) and pixel-level accuracy (P.Acc). We randomly choose a small training set of 16 images and limit data augmentation to horizontal flips to highlight the impact of pre-training, except for Fig.~\ref{fig:trainsize} where we test on different training set partitions. We use Monodepth2 \cite{monodepth2} for depth pre-training. For semantic segmentation, we replace the last layer of the decoder with a fully connected layer, using the finest scale of the multi-scale output. 
We test on ResNet18 and ResNet50 encoders due to their compatibility with various network architectures and widely public-available pre-trained models. Fig.~\ref{fig:diagram} summarizes our experimental setup and Tab.~\ref{tab:results} summarizes the outcomes. In all cases, depth pre-training improves segmentation accuracy.

{\bf Full model. }
Fig.~\ref{fig:comparison} shows the evolution of training loss and model accuracy. Depth pre-training outperforms ImageNet and random initialization. ImageNet pre-training slightly improves over random initialization on ResNet18, but shows almost identical performance to random initialization on ResNet50. Depth also speeds up training, taking $\sim 5000$ iterations to converge, while ImageNet takes 15000 to 20000. Similar results on full-resolution are deferred to the Supp. Mat.

{\bf Different training set size.} Fig.~\ref{fig:trainsize} shows that depth pre-training improves final segmentation scores over all dataset partitions (with ResNet 18).
When training samples increase (\eg 128), ImageNet and depth pre-training (from random initialization) are comparable to random initialization, but depth pre-training initialized with ImageNet yields the best results.

{\bf Different forms of depth supervision.} 
Pre-training quality depends on the source of supervision. Training on monocular videos involves minimizing reprojection error, which requires joint estimation of depth and pose. Since pose estimation relies on sufficiently distinctive textures (large eigenvalues of the structure tensor of image gradients), the supervision signal is sparse. Conversely, with stereo images, one may omit the pose network when training. With depth sensors, training losses minimize error w.r.t. dense or semi-dense measured depth, offering stronger supervision. Tab.~\ref{tab:supervision} shows that supervising with Lidar is the best, followed by stereo, and monocular video -- all improving over ImageNet.

\def\figd{figures/suppmat}
\def\fWidD{0.24\textwidth}
\begin{figure}[t]

\centering
{\footnotesize
\begin{tabular}{c@{\hspace{0.03in}}c@{\hspace{0.03in}}c@{\hspace{0.03in}}c}
\rotatebox{90}{\quad ResNet18}&\includegraphics[width=\fWidD]{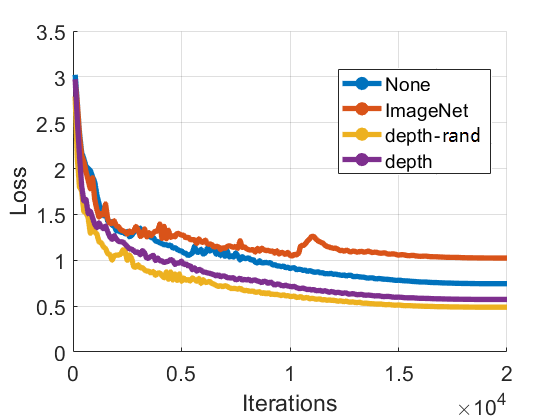} & \includegraphics[width=\fWidD]{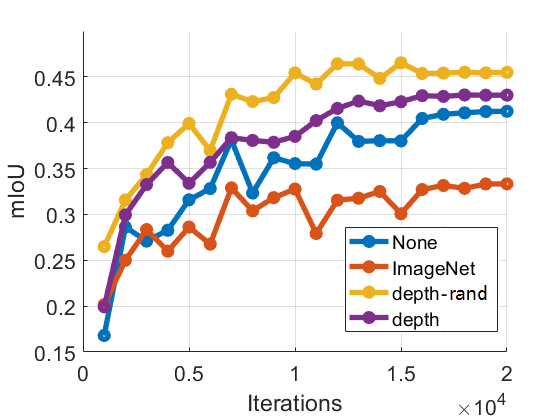} &
\includegraphics[width=\fWidD]{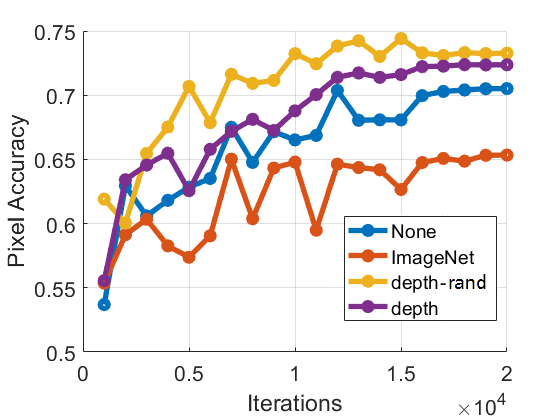} \\
\rotatebox{90}{\quad ResNet50}&\includegraphics[width=\fWidD]{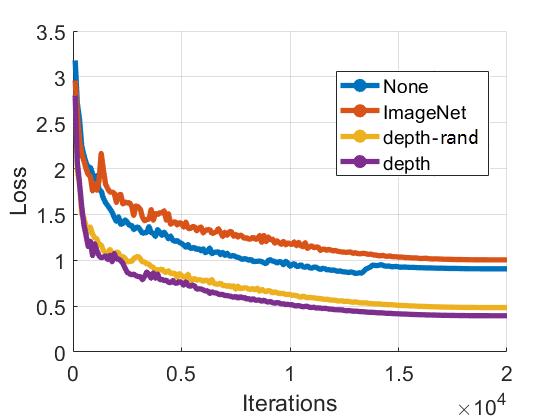} & \includegraphics[width=\fWidD]{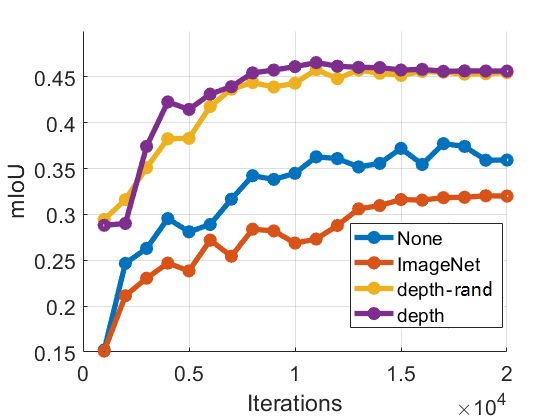} &
\includegraphics[width=\fWidD]{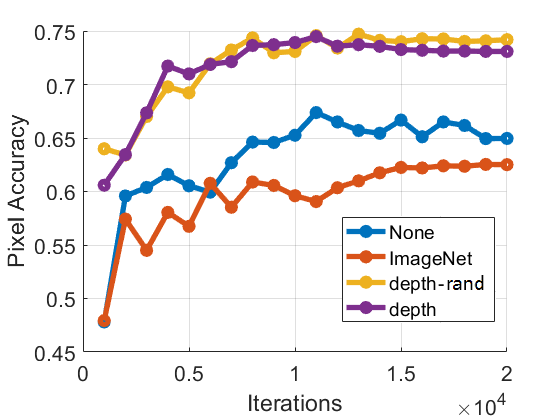}
\end{tabular}
}

\caption{ {\bf Frozen encoder results.} Using an encoder pre-trained by depth significantly outperforms one with random weights and one for ImageNet classification. Note that in this experiment, ImageNet pre-training performs worse than random initialization.}
\label{fig:fixed}
\end{figure}

{\bf Frozen encoder.} We freeze pre-trained encoders, and fine-tune the decoder only, testing the ability of features from pre-trained encoders to capture semantics. With both ResNet18 and ResNet50, encoder pre-trained for depth significantly outperforms random initialization and ImageNet pre-trained (Fig.~\ref{fig:fixed}). It is surprising that ImageNet pre-training is detrimental in this case (after a grid search over learning rates): worse than fixed random weights. This suggests that while classification is a semantic task, it removes semantic information about the {\it scene} due to the object-centric bias in datasets. ImageNet pre-training tends to favor image-level features, that may not capture object shape, making fine-tuning the decoder difficult for segmentation. We conjecture that these uncontrolled biases in ImageNet pre-training cause difficulties in directly predicting segmentation without fine-tuning the encoder. 

\begin{figure}[t]
\begin{minipage}[b]{.45\textwidth}
\centering
 \begin{tabular}{l|l|cc}
&Pre-training& mIoU&P.Acc\\
\hline
\multirow{3}{*}{\rotatebox{90}{Res. 18}}&None&41.35&70.75\\
&ImageNet&45.15&72.39\\
&Depth (encoder only)&\bf{46.69}&\bf{75.04}\\
\hline
\multirow{3}{*}{\rotatebox{90}{Res. 50}}&None&44.66&73.37  \\
&ImageNet&44.65&73.06\\
&Depth (encoder only)&\bf{46.99}&\bf{73.57}\\
\hline
\multirow{2}{*}{\rotatebox{90}{ViT-L}}
&ImageNet&57.53&81.48\\
&Depth (encoder only)&\bf{58.12}&\bf{81.94} 
\end{tabular}
\captionof{table}{ {\bf Initializing with depth encoder and random decoder.} Initializing with the depth encoder and a random decoder outperforms ImageNet initialization, but is worse than initializing with both encoder and decoder from the depth network (see Tab.~\ref{tab:results}).}    \label{tab:fresh_decoder}
\end{minipage}
\hfill
\begin{minipage}[b]{.5\textwidth}
\centering
\includegraphics[width=0.9\linewidth]{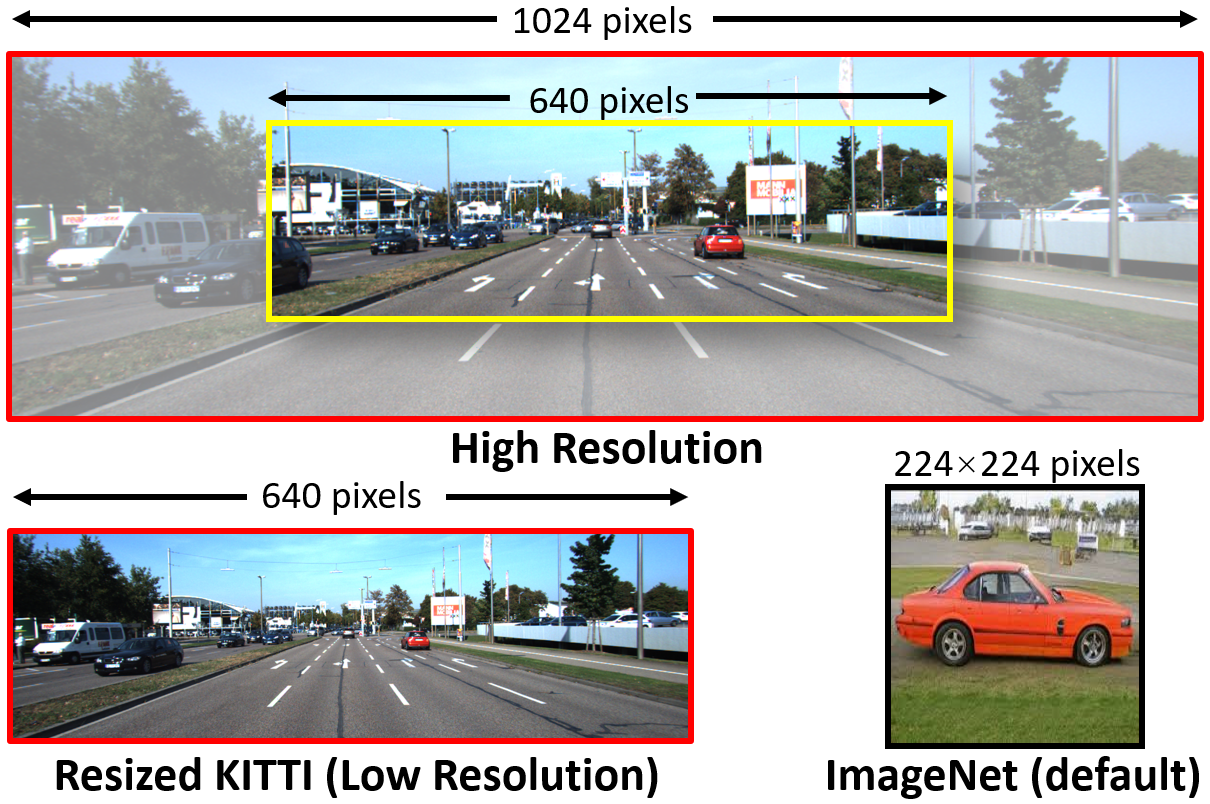}
\caption{ {\bf Mismatch between object scales.} ImageNet models are trained with a fixed input size and objects of interest tend to have a similar scale. However, objects in semantic segmentation dataset vary drastically in scale. Pre-training for depth provides robustness to scale change.}\label{fig:scaling}
\end{minipage}
\end{figure}

{\bf Initializing with a pre-trained encoder only. } 
To eliminate the effect of the depth-initialized decoder and only test the encoder, we replace the decoder by `fresh' randomly initialized weights and fine-tune the whole network. Tab.~\ref{tab:fresh_decoder} shows that depth pre-training outperforms ImageNet when the effect of pre-training is isolate to the encoder. This is also supported by neural activations in Fig.~\ref{fig:activation} where the regions activated after depth pre-training align well with semantic boundaries. Nonetheless, the decoder does play a role in segmentation accuracy: Initializing the whole network with depth pre-training still performs the best, an advantage that is not afforded by a classification head.

\def\figd{figures/suppmat}

{\bf Results on Vision Transformers.}
Given that ViTs~\cite{dosovitskiy2020image} necessitate extensive pre-training, conducting comprehensive ablation studies on ViTs, as feasible with ResNet, becomes impractical due to data constraints. Thus, we conduct experiments using the DPT \cite{ranftl2021vision} depth model trained on a collection of datasets~\cite{lasinger2019towards} (not including KITTI), and report the best results in Tab.~\ref{tab:fresh_decoder} after an exhaustive search for the optimal learning rate. Notably, we identified the optimal learning rate for this experiment to be 5e-8, with larger learning rates yielding suboptimal results. This suggests that DPT inherently provides robust representations sufficient for segmentation, requiring minimal fine-tuning in comparison to CNNs.

{\bf Cross architecture.} To test whether depth pre-training favors our particular choice of architecture, we use the same encoders to initialize DeepLab V3 and follow the same fine-tuning procedure as common practice, \ie, ImageNet initialization. Results are presented in Tab.~\ref{tab:results}. All pre-trainings significantly improve accuracy compared with random initialization. The depth model trained from scratch provides the same level of performance as ImageNet, while subsequent depth training after ImageNet pre-training leads to further improvements.

{\bf Robustness to object scales.}
Given a fixed resolution, ``primary'' objects in object-centric data tend to have a similar scale. For example, ImageNet models are trained on 224$\times$224, thus cars typically have a size of 100 to 200 pixels. In KITTI, however, cars appear at different scales, varying from a few to a few hundred pixels. Fig.~\ref{fig:scaling} illustrates the scale mismatch between datasets. On the other hand, depth pre-training can be done in the same domain as segmentation, hence having robustness to object scales. We examine such robustness: Pre-training on one resolution, fine-tuning on another. Since higher resolution images contain smaller scales (yellow box in Fig.~\ref{fig:scaling}), pre-training on them should still work on smaller images. Pre-trained on smaller resolutions, however, should work poorly on larger images. Our results validate this conjecture: The former achieves a final mIoU of 48.54 (ImageNet: 45.15) while the latter diverges during training.

{\bf Neural Activation.}
We visualize the activations of the ResNet18 encoder by Grad-CAM \cite{selvaraju2017grad}, which was originally designed for classification. We modify it for segmentation by inspecting the gradient of neural response to the summation of predicted labels for pixels instead of one single class label. We visualize shallow layers (before pooling) for high spatial resolution. Fig.~\ref{fig:activation} shows neural activation maps and segmentation outputs from depth pre-training align with semantic boundaries. This confirms not just the similarities between natural and range image statistics discussed in \cite{huang1999statistics}, but also the bias introduced by classification, as activations of ImageNet pre-trained encoder do not resemble object boundaries.

\def\figd{figures/activation}
\def\fWidD{0.24\textwidth}
\begin{figure*}[t]

\centering
{\footnotesize
\begin{tabular}{c@{\hspace{0.005in}}c@{\hspace{0.005in}}c@{\hspace{0.005in}}c}
Image \& GT & Depth & ImageNet & None
\\
\includegraphics[width=\fWidD]{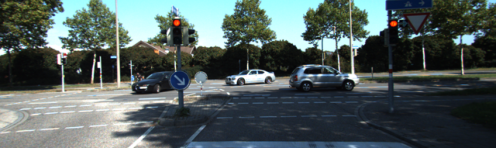}&\includegraphics[width=\fWidD]{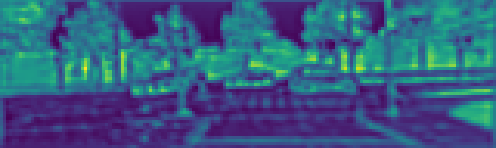} & \includegraphics[width=\fWidD]{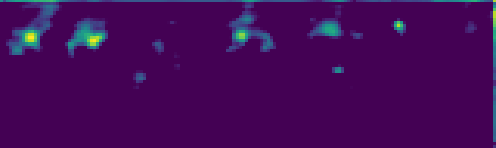} &
\includegraphics[width=\fWidD]{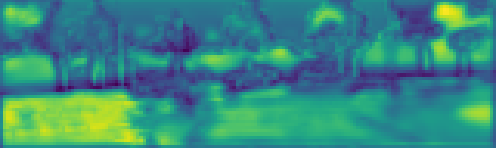} \\[-0.03in]
\includegraphics[width=\fWidD]{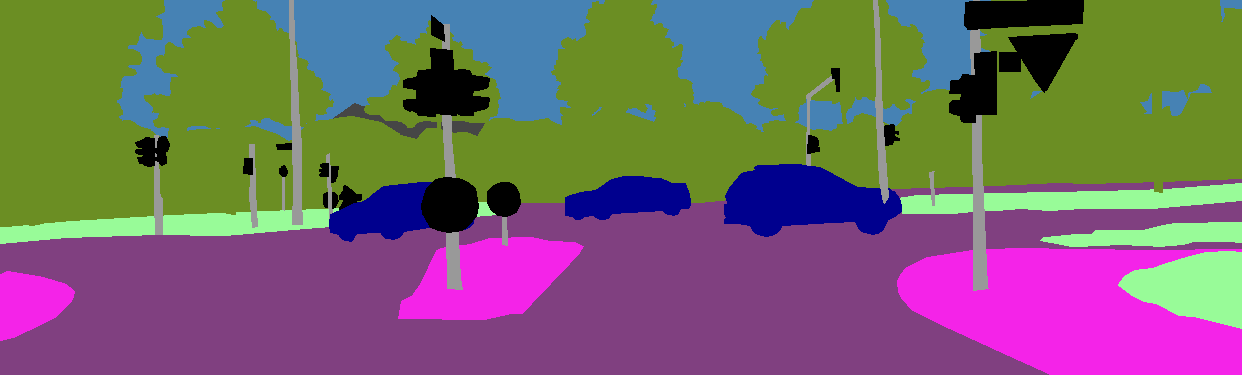}&\includegraphics[width=\fWidD]{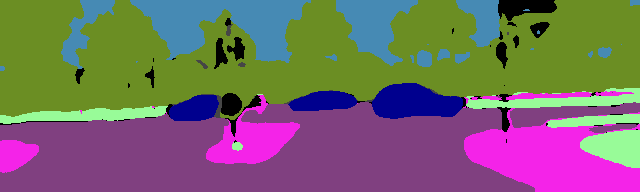} & \includegraphics[width=\fWidD]{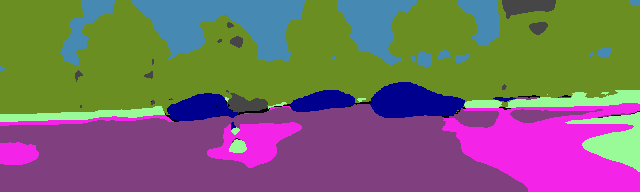} &
\includegraphics[width=\fWidD]{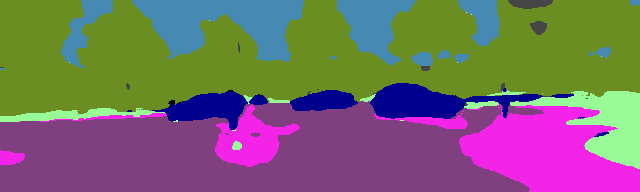}
\end{tabular}
}

\caption{ {\bf Neural activation and semantic segmentation result.} We visualize the neural activation map for a shallow layer of ResNet 18 encoder trained from different initializations, and their corresponding segmentation results. Boundaries are better aligned to semantic boundaries in our model.}

\label{fig:activation}
\end{figure*}

{\bf Comparison with optical flow.}
One hypothesis for the effectiveness of depth pre-training is that the process leverages the statistics of natural scenes where simply-connected components of the range map often correspond to semantically consistent regions. Thus, fine-tuning simply aligns the range of two piece-wise smooth functions. We challenge this hypothesis by trying optical flow, which also exhibits a piecewise-smooth range and is obtained by minimizing the same photometric error. We train optical flow on a siamese network with two shared-weight encoders. While both optical flow and depth capture multiply-connected object boundaries (Fig.~\ref{fig:flow}), using encoders pre-trained for optical flow is detrimental. We conjecture that optical flow does not capture the stable inductive bias afforded by the static component of the underlying scene. Specifically, optical flow is compatible with an underlying 3D geometry only when the scene is rigid, but rigidity is not enforced when inferring optical flow. In contrast, depth forces recognition of rigidity and discards moving objects \cite{lao2017minimum,lao2018extending,lao2023divided} as outliers, which then enables isolating them, also beneficial to semantic segmentation.

\def\figd{figures/flow/}
\def\fWidD{\textwidth}
\begin{figure*}[t]
\centering
{
\begin{minipage}[t]{0.55\textwidth}
\includegraphics[width=0.95\fWidD]{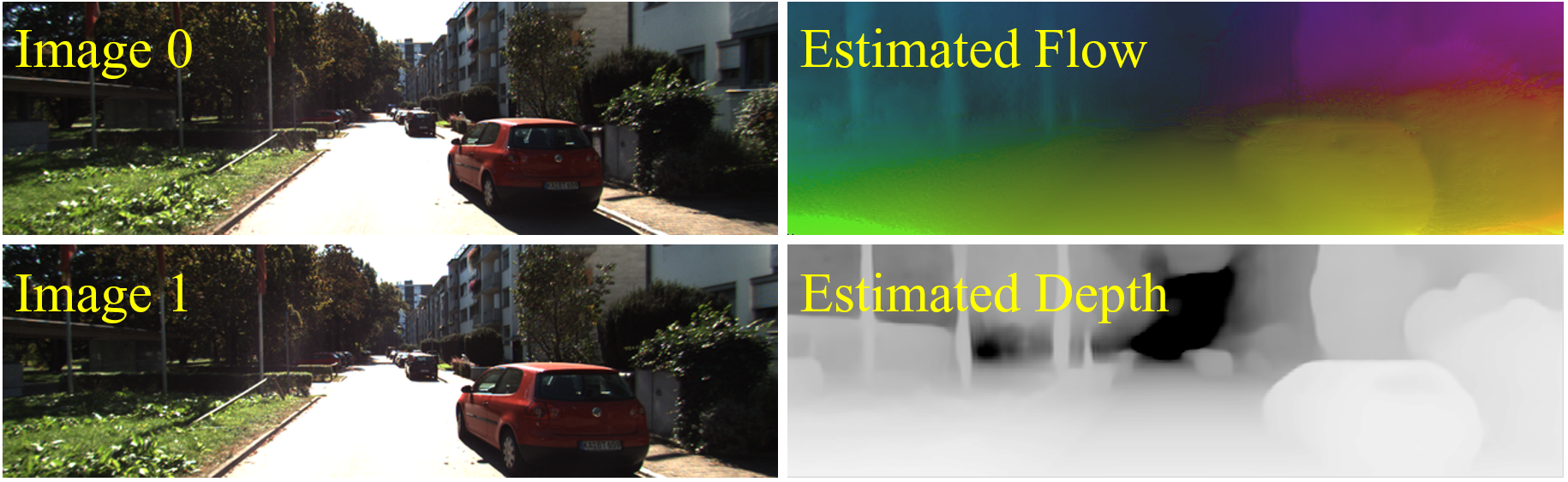}
\end{minipage}
\hfill
\begin{minipage}[t]{0.4\textwidth}
 \footnotesize
\vspace{-0.85in}
\hspace{-0.3in}
  \begin{tabular}{l|cc|cc}
   &\multicolumn{2}{c|}{All}&\multicolumn{2}{c}{Freeze} \\
& mIoU&$\uparrow$&mIoU&$\uparrow$\\
\hline
None&41.35&-&41.24&- \\
Flow&38.47&{\color{red}-2.88}&32.19&{\color{red}-9.05}\\
Depth-Rand&46.00&4.65&43.02&1.78\\
Depth&50.20&8.85&46.53&5.29
  \end{tabular}
\end{minipage}
}

\caption{ {\bf Depth helps, flow hurts.} Although both are pre-trained by minimizing a photometric reconstruction error, monocular depth outperforms optical flow. This stems from the fact that inferring depth from a single image is ill-posed so the network learns inductive priors that are rich in semantics over the structures within a scene. In contrast, any discriminative features will support the correspondence search, so the flow network is not constrained to learning semantics, yielding poor fine-tuning accuracy.
}
\label{fig:flow}
\end{figure*}

\begin{table}[t]
\caption{{\bf Comparison with different pre-trainings.} Encoder: ResNet50; Reconstruction: by inpainting randomly corrupted regions (masked autoencoding); Supervised Segmentation: trained on MS-COCO.}
\setlength{\tabcolsep}{2.7pt} 
 \footnotesize
  \centering
  \begin{tabular}{l|cc|l|cc}
    Pre-training&mIoU&P.Acc&Pre-training&mIoU&P.Acc
    \\
    \hline
    Supervised Segmentation&\bf{51.28}&74.88&Contrastive (DINO)&44.19&71.36\\
    
    Depth&50.92&\bf{77.34}&Optical Flow&42.72&71.80\\
    Reconstruction (MAE) &47.18&74.16& Contrastive (MOCO V2)&37.04&65.91
  \end{tabular}

\label{tab:self_supervised}

\end{table}

{\bf Comparison with other pre-training methods.} In Tab.~\ref{tab:self_supervised}, for completeness, we report results on supervised pre-training by semantic segmentation on MS-COCO \cite{lin2014microsoft}. Unsurprisingly, pre-training on the same task with additional annotated data yields good performance.
Depth estimation, despite not needing additional annotation, yields on-par performance. We also report results on masked autoencoding. We remove random rectangular regions from images, and the network aims to reconstruct the original image. Also known as `inpainting' \cite{bertalmio2000image}, it is considered an effective method for feature learning \cite{pathak2016context}.
It yields inferior performance compared with depth. We conjecture that this is due to artificial rectangular masking which does not respect the natural image statistics, while monocular depth estimation yields occluded regions that border on objects' silhouettes. We also provide results initializing the encoder with contrastive learning (MOCO V2 \cite{chen2020improved} and DINO \cite{caron2021emerging}). While they also remove bias in human annotation for classification, similar to supervised classification, they are still prone to the inherent inductive bias in pre-training data.

\subsection{Full-scale Fine-tuning on the Whole Dataset}

We now extend experiments to fine-tuning on full-scale datasets, aiming to validate the findings drawn from the controlled experiments conducted on KITTI. 

{\bf Cityscapes~\cite{Cordts2016Cityscapes}} contains 2975 training and 500 validation images. Each image has a resolution of $2048\times1024$ densely labeled into 19 semantic classes. The dataset also has 20000 unlabeled stereo pairs with a disparity map, converted to depth via focal length and camera baseline. Like KITTI, Cityscapes is also an outdoor driving dataset. Here we minimize an L1 loss between depth estimates and depth computed from stereo. We modify the prediction head of DeepLabV3 to train for depth, then re-initialize the last layer of the decoder for semantic segmentation. We are unable to reproduce the original numbers. For a fair comparison, we retrained \cite{chen2017rethinking} using the discussed pre-training methods and finetuned under the same training protocol \ie batch size, augmentations, schedule, \etc Tab.~\ref{tab:cityscapes} summarizes the outcomes. We present not only the most favorable results from an extensive training process but also results from a controlled approach with restricted data augmentations and fewer training iterations. Remarkably, improvements remain consistent across both scenarios.

\begin{table}[t]
    \centering
\caption{{\bf Segmentation accuracy on Cityscapes.} Similar to KITTI, pre-training for depth improves segmentation accuracy. Interestingly, under the controlled settings with limited data augmentation and fewer fine-tuning epochs, the model achieves higher performance when pre-training on cropped $256 \times 256$ patches.}
\label{tab:cityscapes}

     \footnotesize
  \begin{tabular}{l|cc|cc|cc|cc}
     &\multicolumn{4}{c|}{Full}&\multicolumn{4}{c}{Controlled}\\
     \hline
   &\multicolumn{2}{c|}{Training}&\multicolumn{2}{c|}{Validation}&\multicolumn{2}{c|}{Training}&\multicolumn{2}{c}{Validation} \\
& mIoU&P.Acc.& mIoU&P.Acc.& mIoU&P.Acc.& mIoU&P.Acc.\\
\hline
None&76.10&95.41&63.97&93.76 &73.82&95.26&60.43&93.07 \\
ImageNet&81.84&96.46&70.41&95.75&76.70&95.71&61.80&93.40\\
Depth&83.46&96.80&\bf{73.17}&\bf{95.24}&77.42&95.82&62.57&93.46\\
Depth-cropped \,\,&\bf{86.80}&\bf{97.41}&72.22&95.01&\bf{79.90}&\bf{96.24}&\bf{65.09}&\bf{94.00}
\end{tabular}

\end{table}
\def\figd{figures/nyu_v2}
\def\fWidD{0.24\textwidth}
\begin{figure}[t]
\centering
 
{\footnotesize
\begin{tabular}{c@{\hspace{0.03in}}c@{\hspace{0.03in}}c@{\hspace{0.03in}}c}
\rotatebox{90}{\quad ResNet18}&\includegraphics[width=\fWidD]{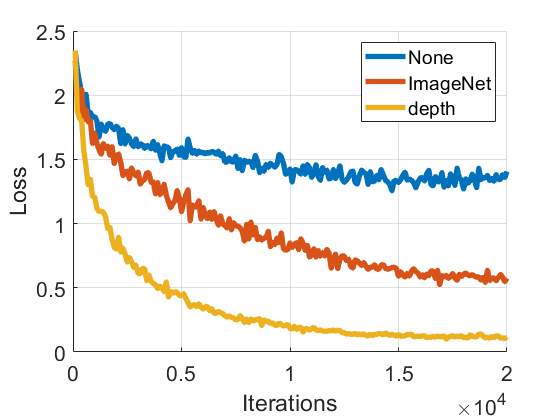} & \includegraphics[width=\fWidD]{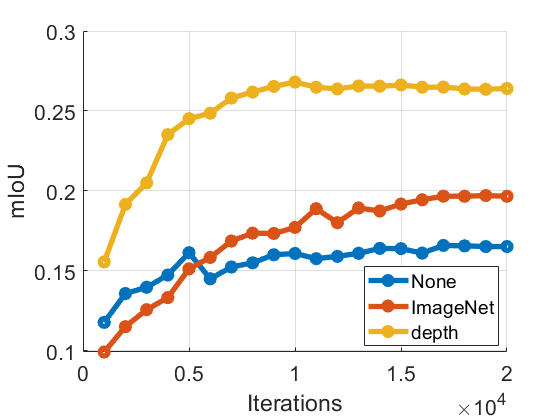} &
\includegraphics[width=\fWidD]{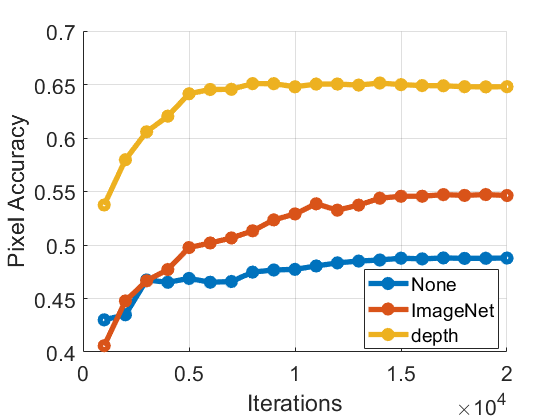} \\
\rotatebox{90}{\quad ResNet50}&\includegraphics[width=\fWidD]{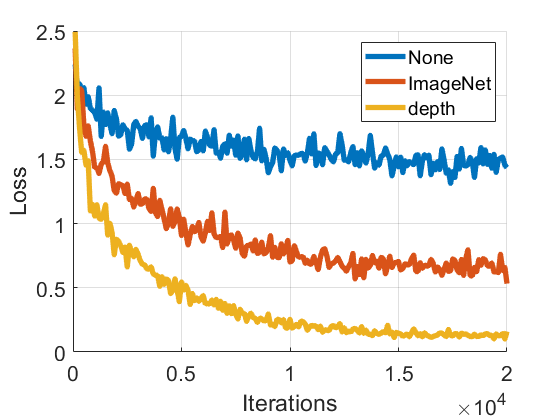} & \includegraphics[width=\fWidD]{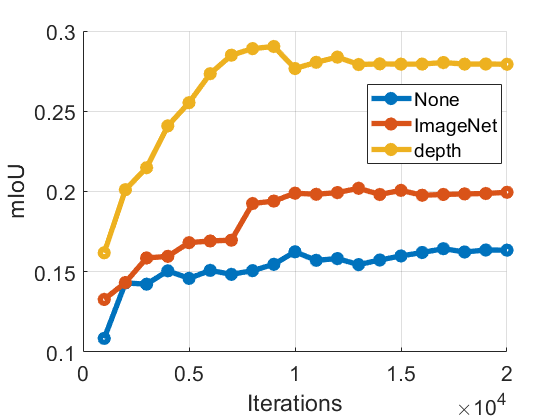} &
\includegraphics[width=\fWidD]{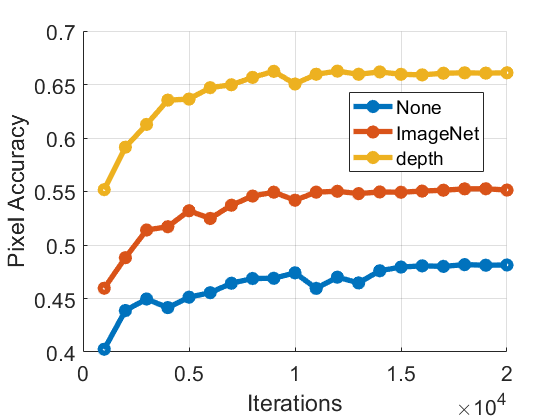}
\end{tabular}
}

\caption{ {\bf Results on NYU-V2.} Similar to KITTI, initializing with depth pre-trained weights trains faster and significantly improves semantic segmentation accuracy.}

\label{fig:nyu}
\end{figure}

One may argue that since depth and semantic segmentation maps are both piece-wise smooth, adapting from depth is naturally easy if the model is aware of each pixel's relative position in the image. In order to test this statement, instead of pre-training for depth on the full image, we train depth on randomly cropped $256 \times 256$ patches, and the model has no spatial awareness of the position of a patch in the image, so depth is purely estimated by local information. This practice (Depth-cropped) surprisingly improves semantic segmentation results under controlled settings, showing that using depth as pre-training goes beyond a simple mapping from one smooth function to another. Interestingly, this approach significantly improves training accuracy in the full setting but leads to a slight reduction in validation accuracy, suggesting a potential issue of overfitting. Future research is necessary to delve into this intriguing phenomenon. Another noteworthy observation is that when pre-trained using depth data, the model exhibits superior performance with a higher initial learning rate of 0.1, as opposed to 0.01 used for ImageNet initialization. Conversely, employing an initial learning rate of 0.1 with ImageNet weights can be detrimental and may result in divergence. These findings suggest that pre-training the network with depth estimation may lead to a smoother local loss landscape.

{\bf NYU-V2~\cite{NYUV2}} is an indoor dataset that contains 795 densely annotated images for training and 654 for testing. There are also 407024 unannotated frames with synchronized depth images captured by a Microsoft Kinect. Since the main hypothesis is agnostic to how depth is learned, we pre-trained for depth using ground-truth as supervision. Unlike outdoor driving, which commonly features sky on top, and road and vehicles in the middle of the image with largely planar camera motion, indoor scenes are characterized by more complex layouts with 6 DoF camera motion, yielding images that are even less likely to resemble the object-centric ones commonly observed in classification datasets. This may be why initializing the model with depth pre-trained weights significantly improves semantic segmentation accuracy with both ResNet18 and ResNet50 (see Fig.~\ref{fig:nyu}). Note that pre-training by depth yields faster convergence, similar to KITTI. 

\subsection{Out-of-domain Transfer from Large-scale Pre-training}

One may conjecture that the advantage of pre-training with monocular depth stems from training within the same domain as downstream semantic segmentation. In practice, in-domain data collection for depth pre-training is indeed ideal since it does not require human labeling. In a scientific context, we are interested in testing this conjecture by investigating the transferability of depth models to tasks outside their original domain. However, common depth datasets are considerably small in scale, making fair comparisons with popular pre-trained models, \eg MAE~\cite{He2021MaskedAA} and DINO v2\cite{oquab2024dinov} that are trained on millions or even billions of images, infeasible. Fortunately, recent methods~\cite{lasinger2019towards, ranftl2021vision} suggest an alternative approach for learning monocular depth by training for relative depth instead of absolute depth, allowing for the integration of multiple data sources during training. Leveraging Depth Anything~\cite{depthanything}, a depth model trained on such scaled-up mixed datasets, we fine-tune for segmentation on three out-of-domain downstream datasets: ADE20k~\cite{ade20k}, PascalVOC~\cite{pascal}, and CityScapes~\cite{Cordts2016Cityscapes}. 

The PascalVOC dataset comprises 10,582 fully annotated images for training purposes and an additional 1,449 for testing, covering a variety of 20 foreground object classes. On the other hand, ADE20k, a sizable dataset, includes 20,210 training images and 2,000 testing images across 150 different classes. The resolution of PascalVOC and ADE20k is $512\times 512$ and $896\times 896$ respectively. On both datasets, we try both fine-tuning the whole network and linear probing with a frozen encoder. Reported in Tab.~\ref{tab:vit_compare}, the results align with the findings on ViTs on the KITTI dataset (in Tab.~\ref{tab:fresh_decoder}), in which case the pre-training data also originates from out-of-domain sources. Notably, consistent improvements in semantic segmentation performance are observed across all datasets compared to the baseline initialization DINO v2~\cite{oquab2024dinov}. It is anticipated that the improvement achieved with a frozen encoder is more substantial than when fine-tuning the entire network, which is consistent with our previous results. This trend underscores that, when trained at a scale, depth models exhibit robust transferability to novel downstream data domains, just as other pre-training methods.

\begin{table}[t]
\caption{{\bf Out-of-domain transfer with large-scale pre-trainings.} We compare Depth Anything with MAE and DINO v2 for semantic segmentation, reporting results (in mIoU) on both fine-tuning (ft) and linear probing (lin.). Depth Anything improves semantic segmentation across all datasets and settings. *: with DINO v2 initialization.}
\setlength{\tabcolsep}{2.7pt} 
 \footnotesize
  \centering
\begin{tabular}{lllcccccccc}
&                                          &  & \multicolumn{2}{c}{ADE20k} &  & \multicolumn{2}{c}{PascalVOC} &  & \multicolumn{2}{c}{CityScapes} \\ \cline{4-5} \cline{7-8} \cline{10-11} 
\multicolumn{1}{l|}{Pre-training}         & \multicolumn{1}{l|}{\# of pre-train data} &  & ft           & lin.         &  & ft             & lin.          &  & ft             & lin.          \\ \hline
\multicolumn{1}{l|}{MAE~\cite{He2021MaskedAA}}            & \multicolumn{1}{l|}{1.28 million}        &  & 53.6         & 49.0         &  & -              & 67.6          &  & -              & 58.4          \\
\multicolumn{1}{l|}{DINO v2~\cite{oquab2024dinov}}        & \multicolumn{1}{l|}{142 million}         &  & 58.1         & 47.7         &  & 86.5           & 86.3          &  & 82.7            & 71.3           \\
\multicolumn{1}{l|}{Depth Anything~\cite{depthanything}} & \multicolumn{1}{l|}{63.5 million*}        &  &\textbf{59.7}       &\textbf{52.3}         &  &\textbf{87.7}           &\textbf{87.3}         &  &\textbf{84.8}           &\textbf{74.8}          
\end{tabular}

\label{tab:vit_compare}

\end{table}

\section{Discussion}
\label{sec:discussion}
Inferring depth only requires multiple images (\eg videos or multiple viewpoints) {\em of the same scene}~\cite{julesz1964binocular} or range sensing, both do not require human-induced priors and bias, unlike semantic tasks that rely entirely on induction: We can associate a label to an image because that image has {\em something} in common with {\em some other} image, portraying a different scene, that some annotator attached a particular label to. That inductive chain has to go through the head of human annotators, who are biased in ways that cannot be easily quantified and controlled. Depth from binocular or motion imagery does not require induction and can be performed {\em ab-ovo}. Learning a monocular model from such supervision eliminates the implicit selective bias from human annotators, yet our findings validate the main hypothesis that the inductive bias learned from such a ``human-free'' process transfers well to the downstream semantic segmentation task. Of course, if different supervisions are available, either for semantic \cite{imagenet_cvpr09}, segmentation \cite{kirillov2023segment}, or both \cite{kuznetsova2020open}, we want to incorporate that information.

Our hypothesis and findings are agnostic to how depth is attributed to a single image: One can perform pre-training using monocular videos, stereo, structure light, LIDAR, or even human guidance \cite{zeng2024wordepth}. One of our pre-training minimizes the photometric reprojection error, used by many predictive and generative approaches. However, an unstructured displacement field is in general not compatible with a rigid motion.  Only if this displacement field has the structure of an epipolar transformation \cite{sundaramoorthi2009set} is the prediction task forced to encode the 3D structure of the scene. This may explain why video prediction is not as effective for pre-training despite many attempts \cite{jin2020exploring,wu2020future,wang2020probabilistic,lao2021flow}.

One limitation of monocular depth estimation is that it may require a calibrated camera, so one cannot use generic videos harvested from the web. There is nothing in principle preventing us from using uncalibrated cameras, simply by adding the calibration matrix $K$ to the nuisance variables. While the necessary conditions for full Euclidean reconstruction are rarely satisfied in consumer videos (for instance, they require cyclo rotation around the optical axis, not just panning and tilting), the degrees of freedom that cannot be reconstructed are moot as they do not affect the reprojection. Moreover, recent progress in training for relative depth from mixed data sources~\cite{lasinger2019towards,ranftl2021vision,depthanything} shows the potential to unlock a virtually unlimited volume of training data, warranting future research.


\bibliographystyle{splncs04}
\bibliography{main_bib}

\clearpage

\appendix
\section{Formalization of Empirical Set-up}
\label{sec:appendixempirical}

The Information Bottleneck Lagrangian in Sect.~\ref{sec:formalization} cannot be computed because we do not have access to the analytical form of the joint distributions of the variables $(\x, \h)$, $(\h, \y)$ and  $(\x, \hat \h)$,  $(\hat \h, \y)$. Even defining, let alone computing, the Shannon Information for random variables that are deterministic maps $\phi(\cdot)$ of variables $\h$ defined in the continuum is non-trivial \cite{achille2019information}. It is possible, however, to bound the Information Bottleneck, which is not computable, with the {\em Information Lagrangian} \cite{achille2018emergence}, which only depends on the datasets ${\cal D}_y, {\cal D}_z$ and ${\cal D}_s$. However, the bound depends on constants that are functions of the complexity of the datasets, which are different for different tasks, which would render them useless in answering the question in \eqref{eq:question}. Therefore, we consider the validation error as a proxy of residual information:\def\x{{\bf{x}}}
\def\y{{\bf{y}}}
\def\z{{\bf{z}}}
\def\h{{\bf{h}}}
\begin{equation}
\hspace{-0.2in}    L_z(w'' | w) = \sum_{\h^n = \phi_w(\x^n)} - \log p_{w''}(\y^n | \h^n) \simeq H(\y | \h)
\end{equation}
for pre-training using depth estimation, and
\begin{equation}
\hspace{-0.2in}     L_y(w'' | w') = \sum_{\hat {\h}^n = \phi_{w'}(\x^n)} - \log p_{w''}(\y^n | {\hat \h}^n) \simeq H(\y | \hat \h)
\end{equation}
for pre-training using another pre-training method. The complexity terms $I(\h; \x)$ and $I(\hat \h; \x)$ are minimized implicitly by the capacity control mechanisms in the architecture ({\em i.e.,} the maps $\phi_w(\cdot | {\cal D}_z)$ and $\phi_{w'}(\cdot | {\cal D}_y)$), for instance pooling; in the regularizers, for instance weight decay and data augmentation  \cite{achille2019information}; and in the optimization, for instance stochastic gradient descent \cite{chaudhari2018stochastic}. The losses above are computed by summing over the samples in the validation set ${\cal D}_s = \{\x^n, \y^n\}$.

\section{Implementation Details}
\subsection{Warping}
\label{sec:appendixwarping}

\begin{equation}
    \hat x(i,j) = x \circ \pi_{g,z}^{-1}(i,j)
   \label{eq:reprojection}
\end{equation}
is the {\em warping} of an image $x$ onto the image plane of another camera related to it by a change of pose $g\in SE(3)$, through the depth map $z$, via a {\em  reprojection map} $\pi^{-1}$ 
\begin{equation}
    \pi^{-1}(i,j) = K_+ \pi (R_t K^{-1} [i,j,1]^T z(i,j)+ T_t)
    \label{eq:z}
\end{equation}
that embeds a pixel $(i,j)$ in homogeneous coordinates, places it in the camera reference frame via a calibration matrix $K$, back-projects it onto the scene by multiplying it by the depth 
$z(i,j) = \phi_w(x(i,j)|{\cal D}_z)$
and then transforming it to the reference frame of another camera with a rigid motion $g = (R, T)$, where the rotation matrix $R\in SO(3)$ and the translation vector $T \in {\mathbb R}^3$ transform the coordinates of spatial points $P\in {\mathbb R}^3$ via $P \mapsto R P + T$. Here $\pi$ is a canonical perspective projection $\pi(P) = [P(1), \ P(2)]/P(3)$ and the calibration map $K_+$ incorporates quantization into the lattice. Here, we assume that the intrinsic calibration matrix $K$ is known, otherwise it can be included among nuisance variables in the optimization along with the inter-frame pose $g_t$ when minimizing the reprojection error $\ell$ in \eqref{eq:reprojection}. 
 
Note that the reprojection error \cite{fei2019geo,wong2019bilateral,wong2020unsupervised,wong2021unsupervised} could be minimized with respect to $w$, which is shared among all images and yields a depth map through $z_t = \phi_w(x_t | {\cal D}_z)$, or directly with respect to $z_t$ in \eqref{eq:z}, which does not require any induction. Since the goal of pre-training is to capture the inductive bias we adopt the former and discuss it in detail in Sect. \ref{sec:discussion}.

\subsection{Training and Evaluation Details on KITTI}
\label{sec:appendixkitti}

\subsubsection{Pre-training for unsupervised depth estimation.}
Monodepth2 is trained by optimizing a linear combination of photometric reprojection error and an edge-aware local smoothness prior
\begin{equation}
    \label{eqn:monodepth_loss}
    L(w') = w_{ph}\ell_{ph}+w_{sm}\ell_{sm},
\end{equation}
where
\begin{align}
    \label{eqn:monodepth_matching}
\ell_{ph}=&\sum_{i, j, n, t} (1-\text{SSIM}(x^n_{t+1}(i,j),\hat{x}^n_{t}(i,j))) \ + \nonumber\\
& \quad\quad\quad \alpha|x^n_{t+1}(i,j)-\hat{x}^n_{t}(i,j)|_1 
\end{align}
$\hat{x}_{t}$ is the warped image \eqref{eq:reprojection} and $\ell_{sm}$ is the edge-aware smoothness prior
\begin{align}
    \label{eqn:smoothness_prior}
\ell_{sm} = \sum_{i, j, n, t} 
      	    &|\partial_{X} z^n_t(i, j)|e^{-|\partial_{X} x^n_t(i, j)|} \ + \nonumber \\
      	    &|\partial_{Y} z^n_t(i, j)|e^{-|\partial_{Y} x^n_t(i, j)|},
\end{align}
$w_{ph}$ and $w_{sm}$ are hyper-parameter weights.

\subsubsection{Semantic segmentation fine-tuning.}
Following the notation in Sect. \ref{sec:formalization} in the main paper, we denote with $\phi_{w''}: \x \mapsto \hat{\y}$ the semantic segmentation network to be fine-tuned, which maps an image $\x$ to a semantic label $\y$. Note that $\phi_{w'}$ (classification network) is parameterized by weights $w'$ of the `encoder' network, and $\phi_{w}$ (depth network) is parameterized by weights $w$ of both the `encoder' and a `decoder'. When pre-trained for classification, we initialize the encoder part of $w''$ by $w'$ and the decoder part by random weights; when pre-trained for depth, we initialize both encoder and decoder in $w''$ using $w$, except for the last layer, where we change to a randomly initialized fully-connected layer with soft-max. During semantic segmentation fine-tuning, we update $w''$ (see equation (11) and (12) in the main paper) by minimizing the cross entropy loss
\begin{equation}
\label{eq:loss2}
    L(w'' |\bullet) = \sum_{i,j,n,k} - \log (\hat{y}^n(i,j)))  \mathbb{1}(y^n(i,j)=k)
\end{equation}
where $i,j$ are the pixel coordinates, $n$ is the number of images in the training set, $k$ is the class label, $\hat{y} = \phi_{w''}(\x^n)$ is the network output. This is implemented via the Negative Log Likelihood (NLL) loss in Pytorch. Under different experimental settings, we either update all parameters in $w''$ or only the decoder part of $w''$ to minimize \eqref{eq:loss2}.

KITTI contains 21 semantic classes, and we fine-tune on all of them. However, since we do not use a separate dataset with segmentation labels, many of the semantic classes are seldom seen ({\em e.g.}, `train', `motorcycle'), especially with just 16 training images. In such cases, these classes always receive zero IoU, which downweights the mIOU metric (but still yields high P.Acc). Therefore, we compute mIoU on a subset of 7 representative classes unless stated otherwise. The results of 21 classes exhibit the same trends.

\subsubsection{Image normalization.}
In fine-tuning, we apply the same image normalization that is consistent with the pre-training step. If the network is pre-trained by ImageNet classification, we normalize the image values by mean=[0.485, 0.456, 0.406], std=[0.229, 0.224, 0.225]; if pre-trained by Monodepth2, we normalize the image values to [0,1]; we also normalize to [0,1] when training from random initialization.

\subsubsection{Optimizer.} After a grid search, we choose $1 \times 10^{-5}$ as the initial learning rate for the ADAM optimizer. The learning rate is updated by a standard cosine learning rate decay schedule in every iteration. 

\subsection{Details for training optical flow}
\label{sec:appendixb}
To train a neural network $f_\theta$ parameterized by $\theta$ to estimate optical flow for a pair of images $(x_t, x_{t+1})$ from time step $t$ to $t+1$, we leverage the photometric reconstruction loss \cite{lao2024diffeomorphic,lao2018extending} by minimizing a color consistency term and a structural consistency term (SSIM) between an image $x_{t+1}$and its reconstruction $\hat{x}_t$ given by the warping $x_t$ with the estimated flow $f_\theta(x_t, x_{t+1})$:

\begin{align}
    \ell_{ph} = \sum_{n, i, j, t}  
    &\lambda_{co} \big(|x^n_{t+1}(i, j) - \hat{x}^n_{t}(i, j)|\big) + \nonumber \\  
    &\lambda_{st} \big(1 - SSIM(x^n_{t+1}(i, j), \hat{x}^n_{t}(i, j))\big),
\end{align}
where $\hat{x}_t = x_t \circ f_\theta(x_t, x_{t+1})$, $f_\theta(\cdot) \in \mathbb{R}^{2HW}$ and $\lambda_{co} = 0.15$, $\lambda_{st} = 0.15$ are the weights for color consistency and SSIM terms for $L_{ph}$, respectively.  

Additionally, we minimize an edge-aware local smoothness regularizer:
\begin{align}
    \ell_{sm} = \lambda_{sm} \sum_{n, i, j} 
      	    &\lambda_{X}(i, j)|\partial_{X} f_\theta(x^n_t, x^n_{t+1})(i, j)| + \nonumber\\
      	    &\lambda_{Y}(i, j)|\partial_{Y} f_\theta(x^n_t, x^n_{t+1})(i, j)|
\end{align}
where $\lambda_{sm} = 5$ is the weight of the smoothness loss, $\partial_{X}, \partial_{Y}$ are gradients along the x and y directions, and the loss for each direction is weighted by $\lambda_{X} := e^{-|\partial_{X}x^n_{t}|}$ and $\lambda_{Y} := e^{-|\partial_{Y}x^n_{t}|}$ respectively.

We trained $f_\theta$ using Adam optimizer with $\beta_1=0.9$ and $\beta_2=0.999$. We set the initial learning rate to be $5 \times 10^{-4}$ for the first 25 epochs and decreased it to $5 \times 10^{-5}$ another 25 epochs. We used a batch size of 8 and resized each image to $640 \times 192$. Because KITTI has two video streams from left and right stereo camera, we randomly sample batches from each stream with a 50\% probability. Training takes $\approx10$ hours for ResNet18 backbone and $\approx20$ hours for ResNet50.

\subsection{Details for training on Cityscapes}
\label{sec:appendixcityscapes}
Different from KITTI, we perform pre-training for depth on a modified DeepLabV3 architecture. We change the prediction head to output a single depth channel instead of 19 classification channels. We follow the standard image normalization of ImageNet training. We use ADAM optimizer with an initial learning rate of 1e-4 and linear learning rate decay. Data augmentations include brightness, contrast, saturation and random horizontal flips. Each model is trained on a single Nvidia GeForce GTX 1080 Ti GPU with a batch size of 8 for 15000 iterations, which takes approximately 6.5 hours. 
During fine-tuning, we re-initialize the prediction head for segmentation. Each model is trained on two Nvidia GeForce GTX 1080 Ti GPUs. We use SGD optimizer with polynomial learning rate decay. We did a learning rate search and report the best performing settings. We find using 0.1 as the initial learning rate yields optimal results when pre-trained by depth, while 0.01 works the best with ImageNet pre-training. All other settings follow the original DeepLabV3 implementation. Limited by the GPU memory we crop training images to $512 \times 512$ patches and set the batch size to be 8. Each model is trained for 60000 iterations (30000 iterations under `controlled' settings), taking \~15 hours, and we report the accuracy after the final epoch.

\subsection{Details for training on NYU-V2}
\label{sec:appendixd}
Because NYU-V2 provide image and depth map pairs, similar to Cityscapes, we directly train Monodepth2 $\phi_w$ by minimizing an $L_1$ loss:
\begin{equation}
    \ell_{L_1} = \sum_{n, i, j}  
    \mathbb{1}(z^n(i,j) > 0) \big(|\phi_w(x^n)(i, j) - z^n(i, j)|\big),
\end{equation}
where $\phi_w(x)$ is the predicted depth for an image $x$ and $z$ is the ground truth depth from a Microsoft Kinect. Because the ground truth is only semi-dense, this loss is only computed where there is valid depth measurements {\em i.e.,} $z(i,j) > 0$. 

We trained Monodepth2 using the ADAM optimizer with $\beta_1=0.9$ and $\beta_2=0.999$. We set the initial learning rate to be $1 \times 10^{-4}$ for 5 epochs and decreased it to $5 \times 10^{-5}$ another 5 epochs for a total of 10 epochs. We used a batch size of 8 and resized each image to $448 \times 384$. For data augmentation, we perform random brightness, contrast and saturation adjustments within the range of $[0.80, 1.20]$ with a 50\% probability. Pre-training the depth network takes $\approx19$ hours for ResNet18 backbone and $\approx30$ hours for ResNet50. After pre-training, we do semantic segmentation fine-tuning on the training set. As with KITTI, we apply the ADAM optimizer with $1 \times 10^{-5}$ initial learning rate and cosine learning rate decay, and restrict data augmentation to horizontal flipping. Limited by GPU memory, for ResNet18 we use batch size 32, and for ResNet50 we use batch size 16. All models are trained for 20000 iterations. Each experiment is repeated by four independent trials.

\section{Experimental Results (Continued)}
\label{sec:experimental_results_cont}

\subsection{Result on high-resolution images (KITTI)}\label{sec:appendixhighreskitti}
In Fig.~\ref{fig:highres_resnet} we show results trained with full resolution (1024$\times$320) and full dataset (200 images), in which case we train for an extended number of iterations (50000). Pre-training by depth still outperforms ImageNet pre-training and random initialization. Note that in this case using random initialization is unstable and training diverges after 30000 iterations. 

\def\figd{figures/suppmat}
\def\fWidD{0.22\textwidth}
\begin{figure}[t]
\centering
\hspace*{-0.25in}
{\footnotesize
\begin{tabular}{c@{\hspace{0.03in}}c@{\hspace{0.03in}}c@{\hspace{0.03in}}c}
\rotatebox{90}{\quad\quad ResNet18}&\includegraphics[width=\fWidD]{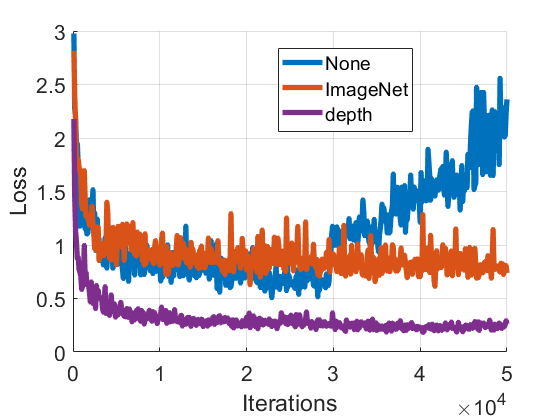} & \includegraphics[width=\fWidD]{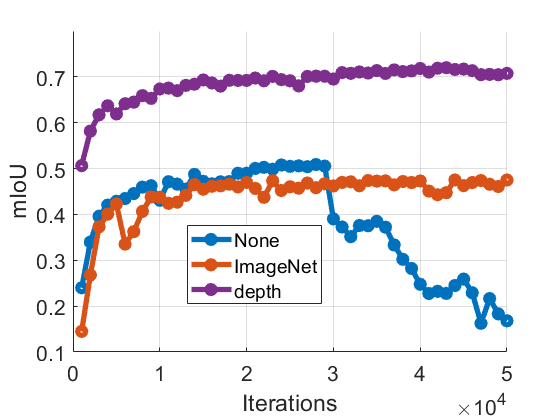} &
\includegraphics[width=\fWidD]{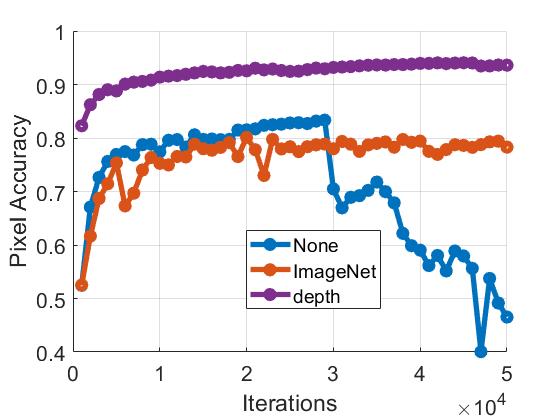}
\end{tabular}
}
\caption{\sl\small {\bf Training ResNet with high resolution images.} Similar to low resolution images, pre-training on depth also improves semantic segmentation accuracy on high resolution images.}

\label{fig:highres_resnet}
\end{figure}

We also conduct experiments on high-resolution images. Results are consistent with low-resolution experiments. Although both ImageNet and depth initialization converged to the same level of accuracy, the depth pre-trained model shows a high accuracy in earlier iterations. This is interesting given that training loss of both ImageNet and depth pre-train are almost the same. We conjecture that the features learned by single-image depth estimation are more conducive to segmenting `bigger' classes (e.g. `road', `building') which are mostly rigid and take up larger portions of the image.

To further investigate this behavior, in Figure \ref{fig:miou_pix} we plot the mIoU curve on all (21 classes), and a scatter plot of mIoU versus pixel accuracy in the training process. While mIoU on all 21 classes follows the trend observed on 7 classes, the scatter plot shows that at each same level of mIoU, models pre-trained by depth have higher pixel accuracy. This validates our conjecture that the depth-pre-trained model learns `bigger' classes faster, since higher performance on these classes will result in high pixel accuracy as they have more pixels.

\def\figd{figures/suppmat}
\def\fWidD{0.22\textwidth}
\begin{figure}[t]
\centering
\hspace*{-0.25in}
{\footnotesize
\begin{tabular}{c@{\hspace{0.03in}}c@{\hspace{0.03in}}c@{\hspace{0.03in}}c}
\rotatebox{90}{\quad\quad Deeplab}&\includegraphics[width=\fWidD]{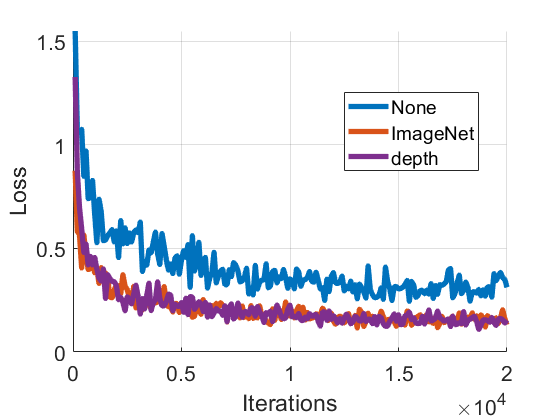} & \includegraphics[width=\fWidD]{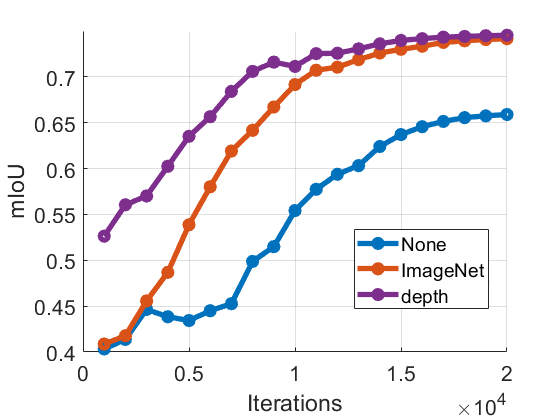} &
\includegraphics[width=\fWidD]{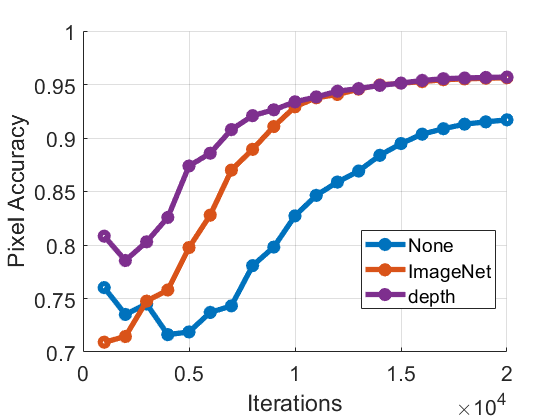}
\end{tabular}
}
\caption{\sl\small {\bf Training Deeplab on high resolution images.} Compared to lower resolutions, performance improves for different initializations. Training loss (left) is similar between ImageNet and depth initialization, but mIOU (center) and pixel accuracy (right) are higher for depth initialization for similar loss values.}
\label{fig:highres_deeplab}
\end{figure}

\def\figd{figures/suppmat}
\def\fWidD{0.22\textwidth}
\begin{figure}[t]
\centering
{\footnotesize
\begin{tabular}{c@{\hspace{0.05in}}c}
\includegraphics[width=\fWidD]{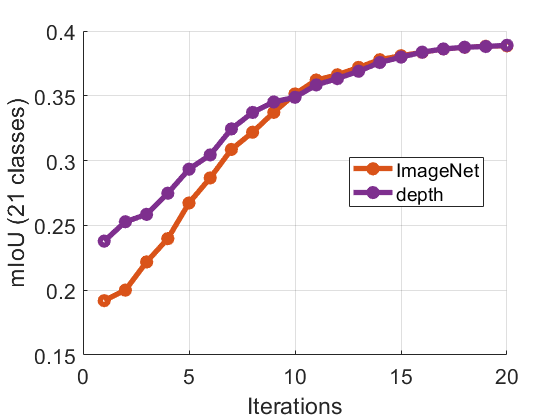}
&\includegraphics[width=\fWidD]{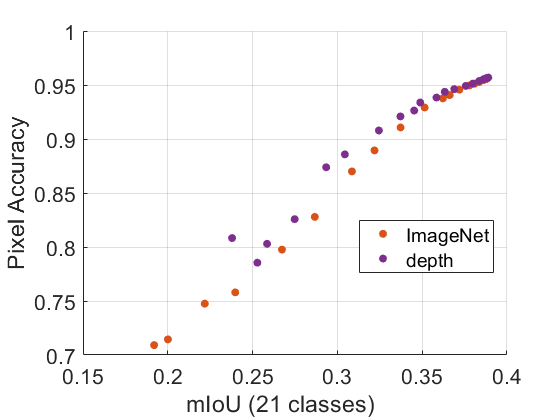} 
\end{tabular}
}

\caption{\sl\small {\bf mIoU v.s. pixel accuracy during training DeepLab.} At each same level of mIoU, model pre-trained by depth has a higher pixel accuracy. This validates our conjecture that the depth-pre-trained model learns `bigger' classes faster.}
\label{fig:miou_pix}
\end{figure}

\subsection{Training loss on Cityscapes}

\def\figd{figures/suppmat}
\def\fWidD{0.3\textwidth}
\begin{figure}[t]
\centering
{\footnotesize
\includegraphics[width=\fWidD]{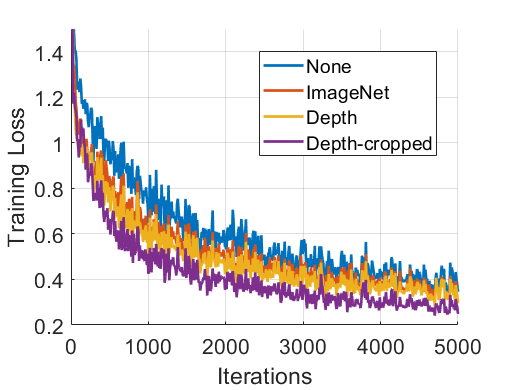}
}

\caption{\sl\small {\bf Training loss of early iterations on Cityscapes.} In early training iterations, the loss of {\it Depth-cropped} decreases significantly faster than other initializations.}
\label{fig:loss_cityscapes}
\end{figure}

In Fig.~\ref{fig:loss_cityscapes} we show training loss of early iterations on Cityscapes shows that training on small crops of the image improves convergence. We revisit the question we posed in Table 4 in the main text regarding this phenomenon below.

\subsection{Visualizations of Segmentation}
\cref{fig:cityscapes_head_to_head} shows head-to-head comparisons of representative outputs from semantic segmentation models that have been pre-trained with ImageNet classification, and pre-trained with monocular depth estimation. As discussed in Sect. \ref{sec:setups} Neural Activations, pre-training on ImageNet biases the model towards capturing generic textures exhibited in the image rather than that object shape (see Fig.~\ref{fig:activation}). This results in a loss of details when fine-tuning for the downstream segmentation task where the goal is precisely to capture object boundaries. We illustrate the drawback of pre-training on ImageNet in \cref{fig:cityscapes_head_to_head}-left where the model over-predicts the street sign in the center (highlighted in yellow) and under-predicts the pedestrians on the right (highlighted in red) with spurious predictions of the vehicle class alongside them. This is in contrast to pre-training with monocular depth estimation, where the downstream segmentation model is able to capture the edge between the street sign and building regions in the middle as well as the small pedestrian regions in the far distance on the right. Additionally, we show in \cref{fig:cityscapes_head_to_head}-right that an ImageNet pre-trained model has difficulty outputting the same class for a consistent surface like the sidewalk on the left, which is unlike a model pre-trained on depth. This is also supported by our results in \cref{sec:experimental_results_cont} and Sect. \ref{sec:setups} in the main text where the ``larger'' object classes tend to be more easily learned (higher general P.Acc) by a model pre-trained on depth than one trained on ImageNet. 

While classification is a semantic task, training for it requires discarding nuisances including objects (other than one that is front and center, see neural activations in Fig.~\ref{fig:activation} and results with a frozen pre-trained encoder in Fig.~\ref{fig:fixed}) that may exist in the background of an image. Pre-training for depth, on the other hand, involves solving correspondence problems, which naturally requires estimating the boundaries of objects and consistent locally-connected, piecewise smooth surfaces. Thus, one may hypothesize that pre-training to predict depth or geometry process makes it more straightforward to assign these surfaces with a semantic class or label i.e. road, pedestrian, building. We conjecture that this may play a role in sample complexity as illustrated by the performance improvements observed over ImageNet pre-training when training with fewer samples (Fig.~\ref{fig:trainsize}).

\def\figd{figures/suppmat/cityscapes}
\def\fWidD{0.23\textwidth}
\begin{figure}[t]
\hspace*{-0.1in}
\centering
{\footnotesize
\begin{tabular}{c@{\hspace{0.01in}}c@{\hspace{0.005in}}c}
\rotatebox{90}{\quad\quad Image}&\includegraphics[width=\fWidD]{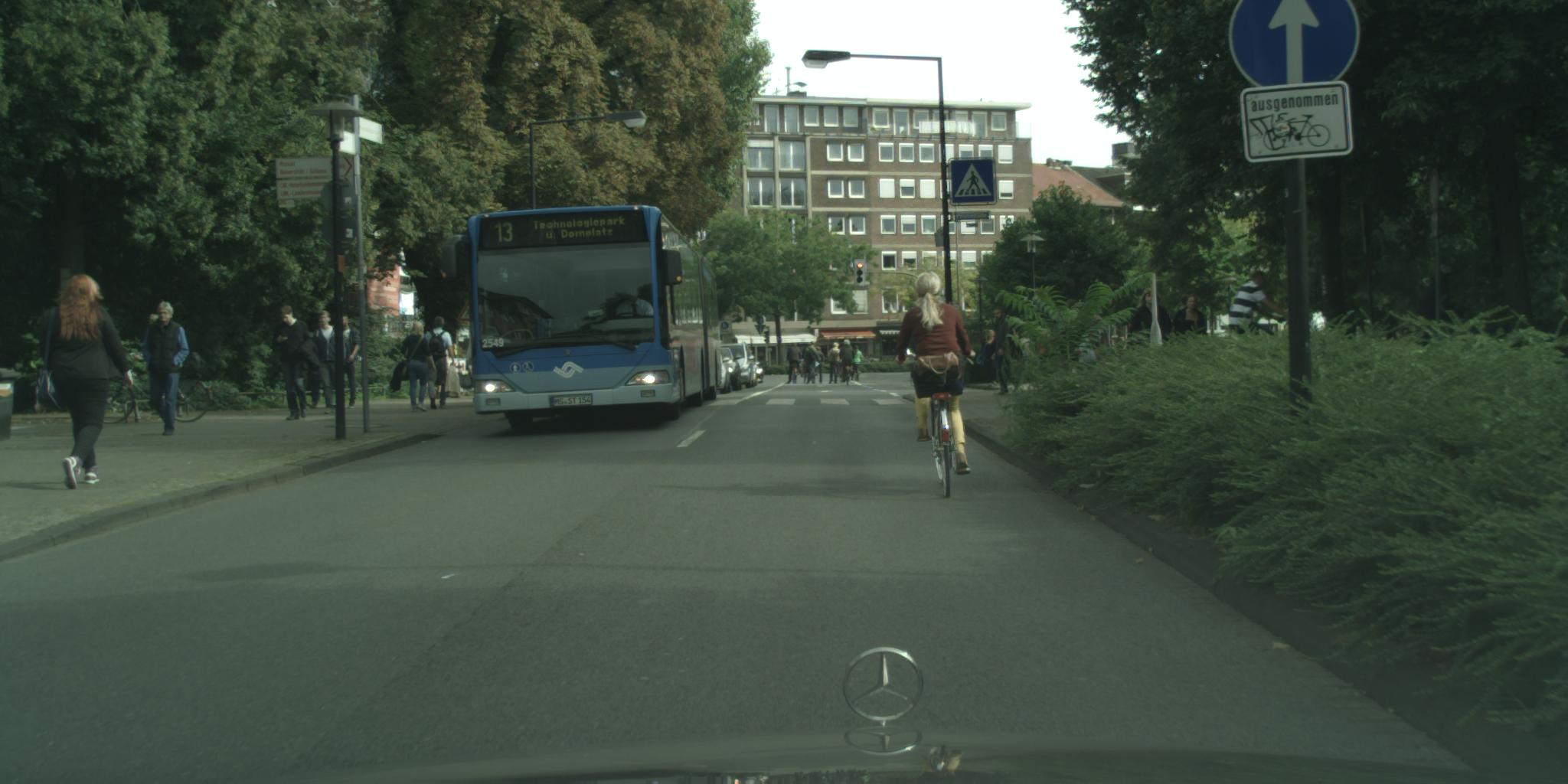}&\includegraphics[width=\fWidD]{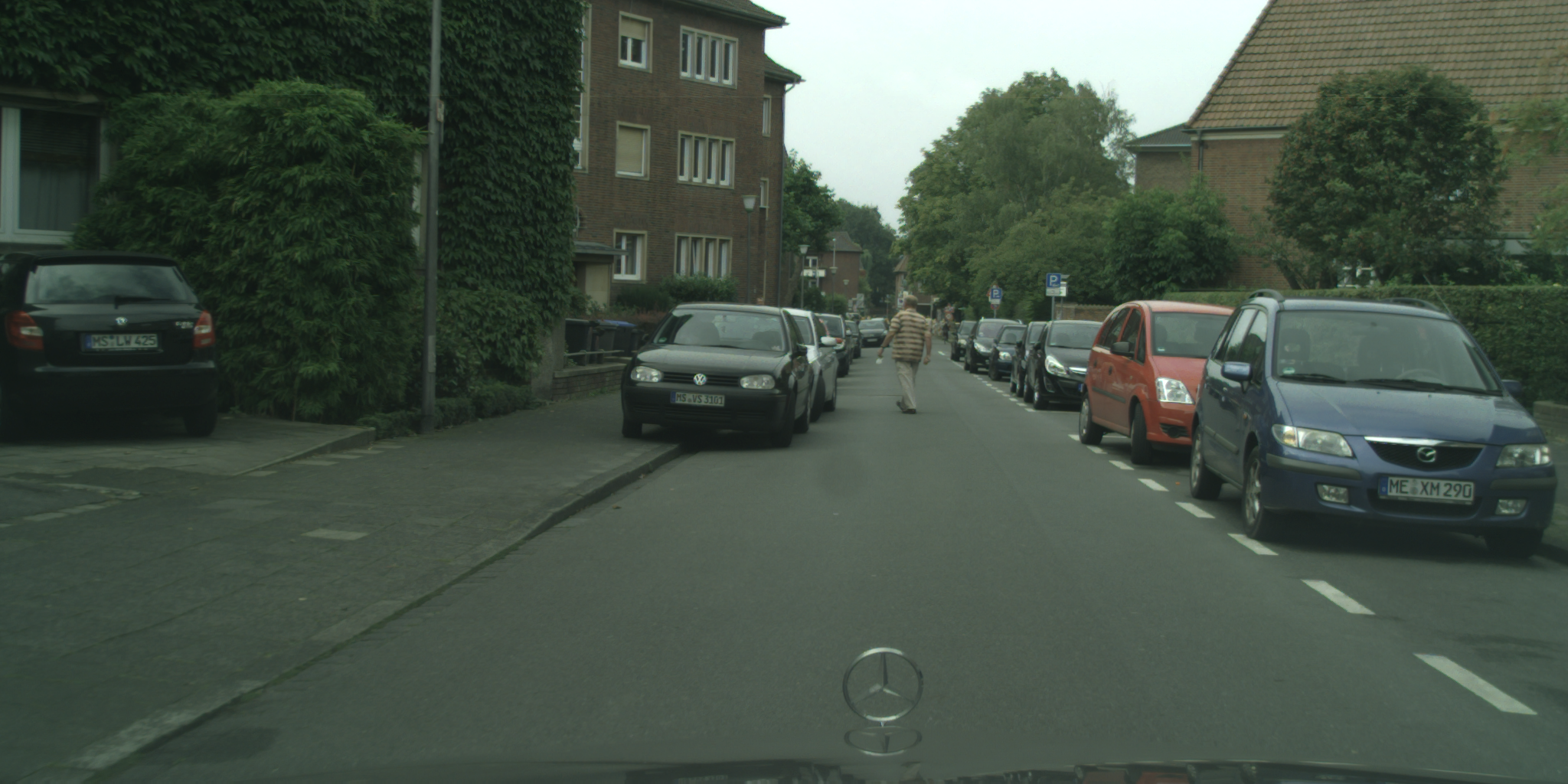}\\[-0.03in]
\rotatebox{90}{\quad Groundtruth}&\includegraphics[width=\fWidD]{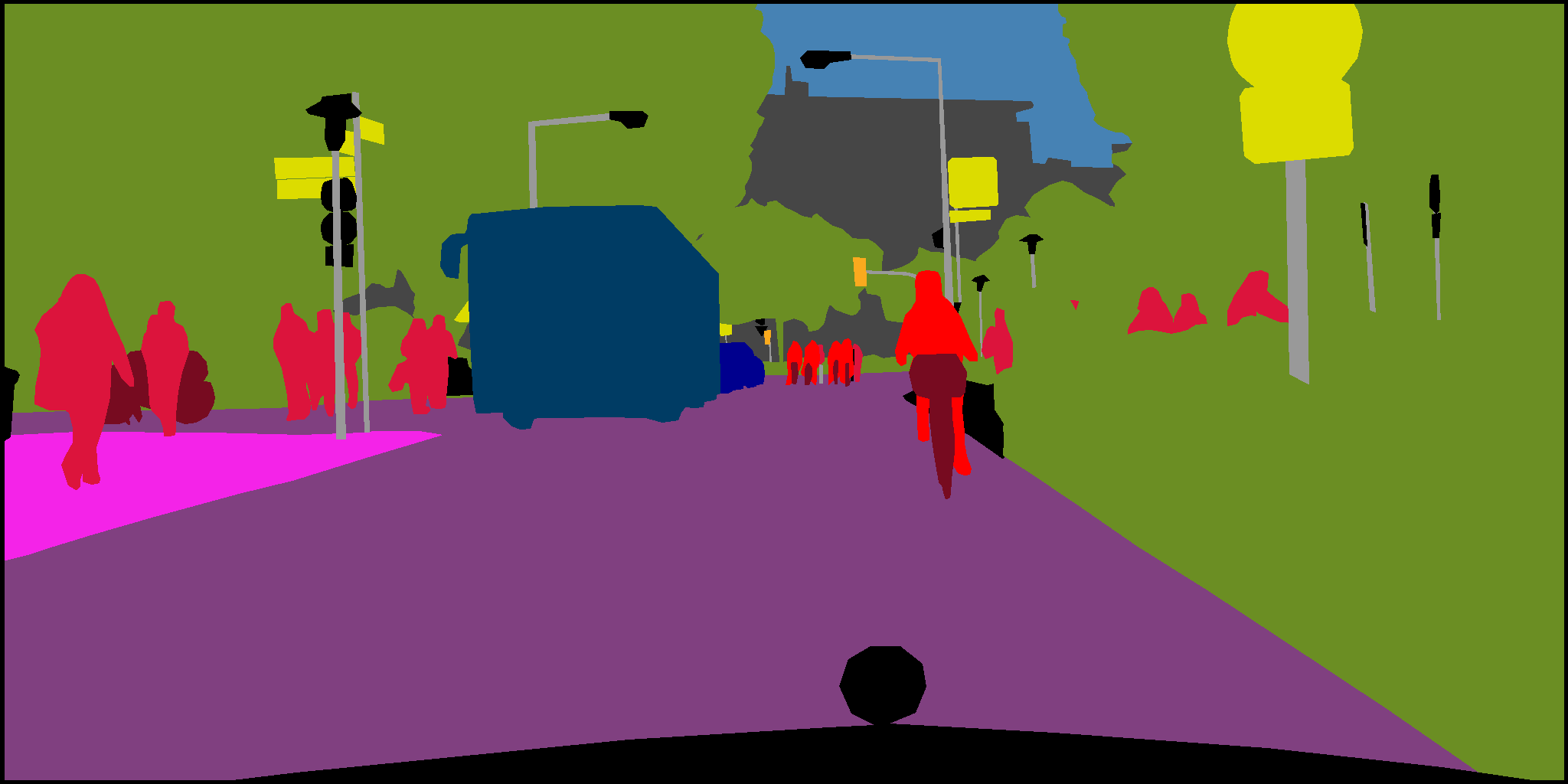}&\includegraphics[width=\fWidD]{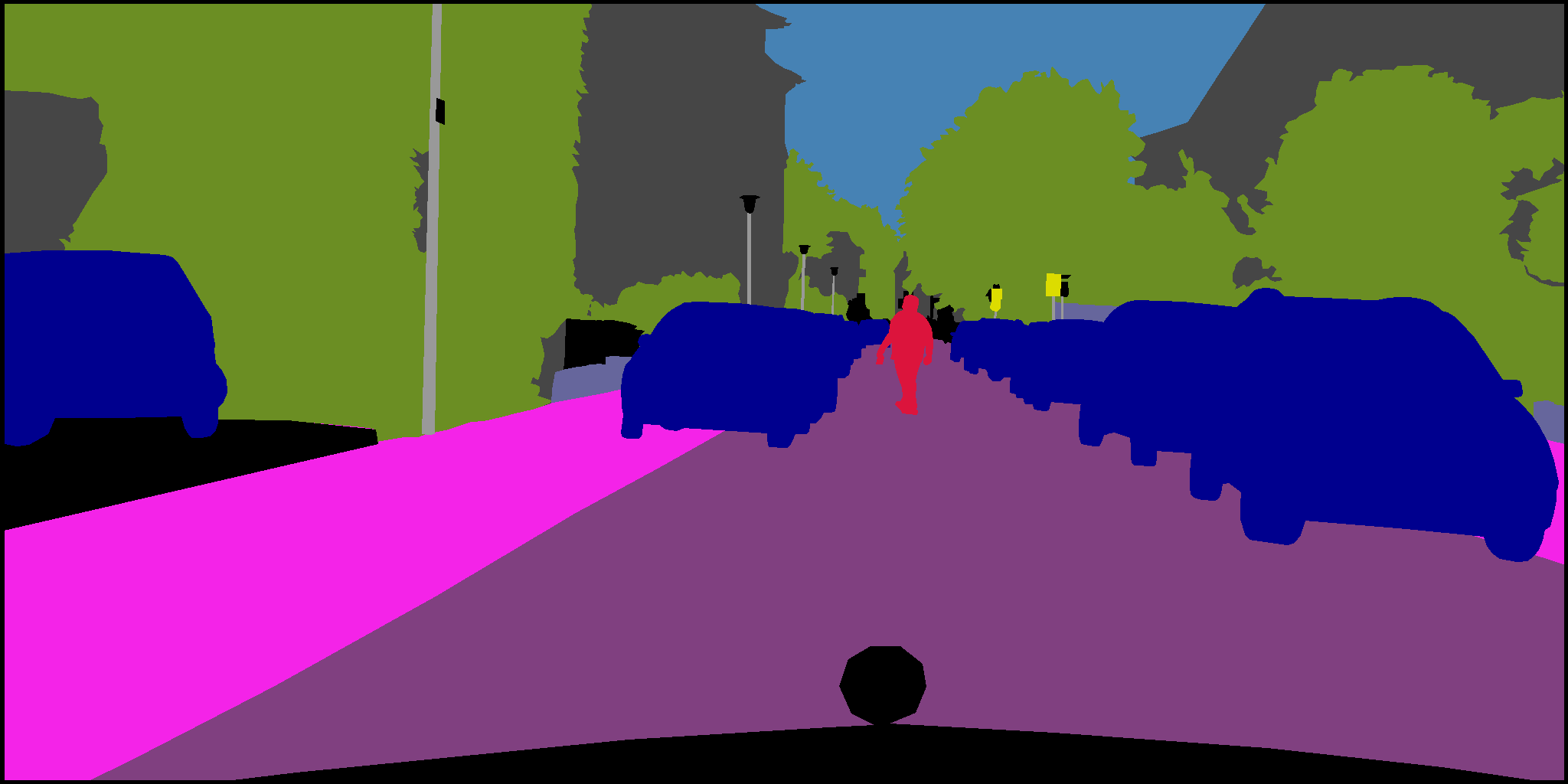}\\[-0.03in]
\rotatebox{90}{\quad\quad Ours}&\includegraphics[width=\fWidD]{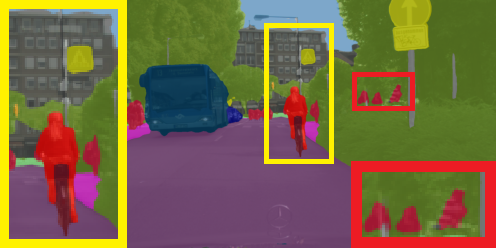}&\includegraphics[width=\fWidD]{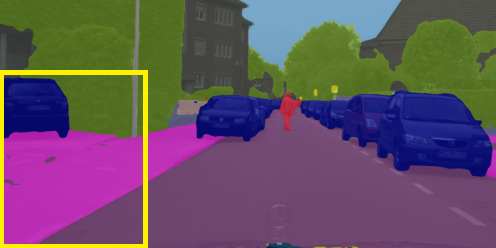}\\[-0.03in]
\rotatebox{90}{\quad\quad ImageNet}&\includegraphics[width=\fWidD]{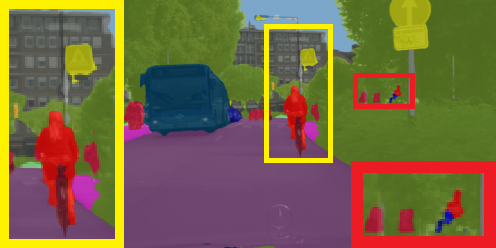}&\includegraphics[width=\fWidD]{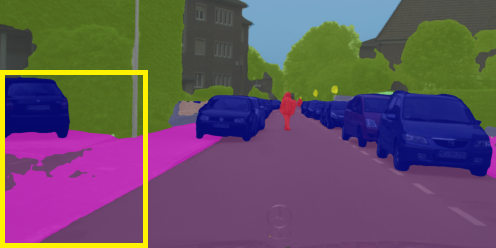}
\end{tabular}
}
\vspace{-0.6em}
\caption{\sl\small {\bf Head-to-head comparison between ImageNet and monocular depth estimation pre-training for semantic segmentation on Cityscapes.} We visualize the representative outputs of segmentation models pre-trained on monocular depth (row 3) and ImageNet (row 4). Pre-training with depth enables sharper object boundaries, i.e. street sign highlighted in yellow, pedestrians highlighted in red, and more consistent class predictions on large objects like the sidewalk on the left in the right column. We note that pre-training on ImageNet also yields spurious predictions like the vehicle class next to the pedestrians (highlighted in red) and the street sign class on the street light in the middle of the left image. Better viewed zoomed-in and in color.}
\label{fig:cityscapes_head_to_head}
\end{figure}

\subsection{Additional Results on Object Detection}

To demonstrate the versatility of depth pretraining, we extend our experiments to object detection. Using DepthAnything as the depth pre-training method, we conducted a comparison with DINO-V2. The results are shown in Tab.~\ref{tab:object_detection}. Our findings reveal that depth as pre-training outperforms DINO-V2 on the ADE20K and Cityscapes datasets, while yielding comparable performance on COCO. This outcome is anticipated because DINO-V2's pre-training dataset is object-centric, aligning closely with COCO's characteristics. In contrast, ADE20K and Cityscapes datasets exhibit a bigger domain gap, suggesting that depth pre-training may generalize to diverse data domains. We note that this experiment only marks the tip of the iceberg; related tasks include moving object detection \cite{lao2017minimum,lao2019minimum}, which we leave for future work.

\begin{table}[h!]
  \centering
      \caption{\sl\small {\bf Quantitative results of object detection.} We initialize a Vit-L for MaskRCNN using DINOv2 and DepthAnything pretrained weights on three object detection datasets. Depth pretraining improves on ADE20K and Cityscapes while being comparable on COCO.}
    \vspace{-3mm}
    \scalebox{0.85}{
    \begin{tabular}{l c c c c c c c c c}
        \midrule
         & \multicolumn{3}{c}{ADE20K} & \multicolumn{3}{c}{Cityscapes} & \multicolumn{3}{c}{COCO} \\
         \cmidrule(lr){2-4} \cmidrule(lr){5-7} \cmidrule(lr){8-10}
         Pretraining & mAP$\uparrow$ & mAP@50$\uparrow$ & mAP@75$\uparrow$ & mAP$\uparrow$& mAP@50$\uparrow$& mAP@75$\uparrow$ & mAP$\uparrow$& mAP@50$\uparrow$& mAP@75$\uparrow$ \\ 
        \midrule
        DinoV2
        & 0.326 & 0.509 & 0.354 & 0.321 & 0.540 & 0.310 & \textbf{0.499} & 0.719 & \textbf{0.540} \\
        \midrule
        Depth
        & \textbf{0.334} & \textbf{0.521} & \textbf{0.355} & \textbf{0.327} & \textbf{0.548} & \textbf{0.332} & \textbf{0.499}	& \textbf{0.720} & 0.539 \\
        \midrule
    \end{tabular}}
    \vspace{-1em}
\label{tab:object_detection}
\end{table}

\section{Extended Discussion}
\label{sec:appendixdiscussion}
\subsection{Cropping Size During Pre-training}

One may object that once a depth map has been estimated, a semantic map is just a matter of aligning labels. However, depth is not necessarily a piecewise constant function, although it is generally piecewise smooth. So, for instance, the road at the bottom center of KITTI images is a slanted plan that does not correspond well to any constant value, yet the model converts it into a consistent class.

One may also object that slanted planes at the bottom center of the visual field have a strong bias towards being labeled ``road'' given the data on which the model is trained. 

For this reason, we conducted the experiments shown in Tab.~\ref{tab:cityscapes}, whereby we select random crops of 3\% of the size of the image, and use those for pre-training rather than the full image. This way, there is no knowledge of the location of the patch of the slanted plane relative to the image frame. We were expecting a degradation in performance, but instead observing an improvement.

While in theory full consideration of the visual field is more informative, provided sufficient training data, due to the limited volume of the training set and the strong biases in the training data, breaking the image into smaller patches and discarding their relation (position on the image plane) may help break the spurious dataset-dependent correlations and lead to better generalization after fine-tuning.

This conclusion is speculative, and we leave full testing to future work. The important aspect of this experiment is to verify that fine-tuning semantic segmentation after depth pre-training is not just a matter of renaming a piecewise smooth depth field into a piecewise constant labeled field. 

\subsection{Relation to Prior Arts}\label{sec:appendixpriorarts}
The primary objective of this study is to conduct an exhaustive examination of the adoption of monocular depth as a pre-training technique for semantic segmentation and to compare it with classification (which has been the de-facto approach for weight initialization). It is acknowledged that prior research \cite{hoyer2021three,hoyer2021improving,jiang2018self} has also recommended the use of depth as a pre-training method. While we are not claiming originality in using depth, our study stands out as the first to systematically investigate this issue under different setups including network architecture, resolution, and supervision. Nevertheless, there are also distinctions between our study and these methods. 

\cite{jiang2018self} performs pre-training by estimating relative depth on images, while we focus on estimating absolute depth either through supervised or unsupervised pre-training. We present results that contradict those of \cite{jiang2018self}. While \cite{jiang2018self} claim that ImageNet is the better pre-training approach, our findings show better performance when using our depth pretraining as compared to ImageNet initialization alone. The difference in findings is suprising as we observe consistent performance boost across our experiments; one possible reason may be that camera calibration is available to us (also a valid assumption for real-world scenarios that require semantic segmentation, such as self-driving), and it is well-known that un-calibrated depth estimation is a more difficult problem than the calibrated case. 
Another difference (and the possible reason for the contradicting results) is that \cite{jiang2018self} synthesizes relative depth from optical flow estimation network, and train a depth prediction network by minimizing L1 difference between predictions and relative depth. The authors also state that potential errors in optical flow will impact (and compound the error in) the downstream depth. 

\cite{hoyer2021three, hoyer2021improving} use self-supervised depth prediction as a proxy task for semantic segmentation, where the features are regularized to stay close to ImageNet features. However, our experiments demonstrate that in some cases, ImageNet features may not support the task of semantic segmentation as observed in Fig.~\ref{fig:activation}, and also empirically validated in \cref{fig:cityscapes_head_to_head}. Even when we train the depth model with ImageNet initialization, we do not assume that the features should remain largely unchanged. Nonetheless, we are not against using ImageNet, as it is off-the-shelf and can be useful to expedite training depth estimation networks, likely due to the formation of generic filter banks in the early layers.

We would emphasize that our paper validates the findings in prior works \cite{jiang2018self,hoyer2021three,hoyer2021improving}, yet provides a more comprehensive longitudinal study (across different architectures, datasets, supervision, etc.) towards using monocular depth as pre-training for semantic segmentation.

\subsection{Insights On the Intuitions and Feasibility of Using Monocular Depth as Pre-training} \label{sec:appendixintuitions}
\begin{figure*}[t]
\centering
\includegraphics[width=0.75\textwidth]{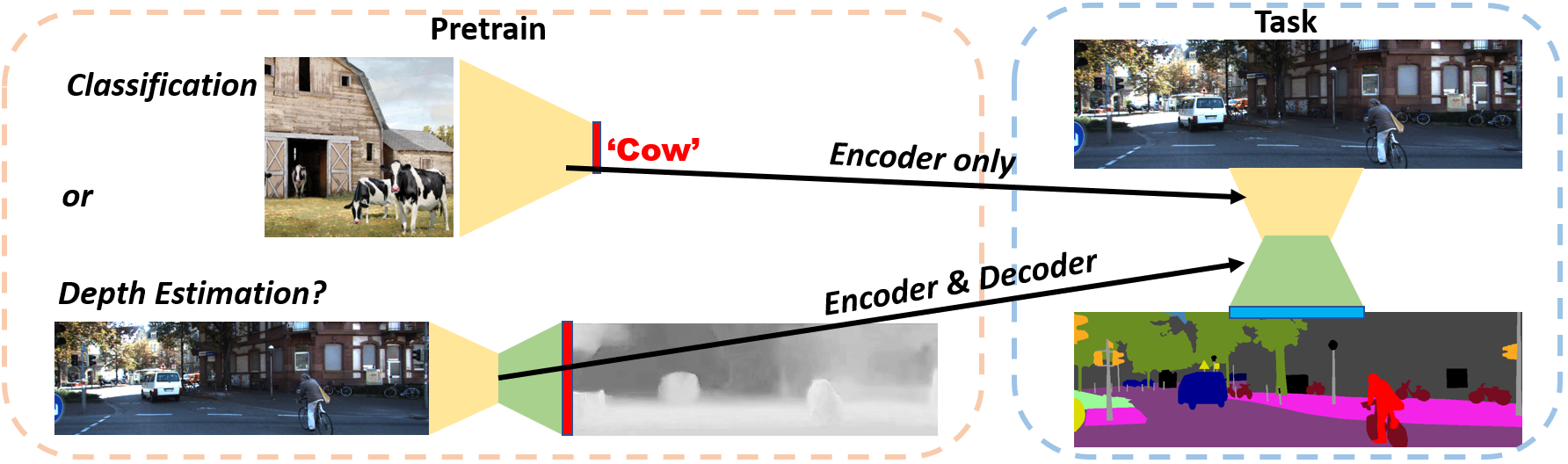}
  \caption{\sl\small Image classification introduces uncontrolled biases when used to pre-train semantic segmentation networks, where one requires an additional decoder. Depth estimation, on the other hand, is not subject to semantic bias in the pre-training dataset, and eliminates the need for human annotation, and can easily adapt the pre-training dataset to the domain of interest. The question is whether such a process can improve performance and reduce dependency of annotated pre-training datasets in fine-tuning for semantic segmentation.}
 \label{fig:intro}
\vspace{-1em}
\end{figure*}

ImageNet classification is widely regarded as the primary task for pre-training in the context of semantic segmentation. The dataset comprises more than 14 million annotated images, collected through crowd-sourcing with the assistance of around 15,000 paid annotators. Empirical studies have consistently confirmed the advantages of using ImageNet for initial model training, resulting in substantial enhancements in the performance of semantic segmentation tasks. This outcome is unsurprising, as both image classification and semantic segmentation involve understanding the meaning of objects in images. It's important to note that obtaining detailed pixel-level annotations for semantic segmentation is a costly and resource-intensive process. Consequently, datasets specifically tailored for semantic segmentation are considerably smaller in scale compared to ImageNet, often differing by several orders of magnitude in terms of data size.

In contrast, the KITTI dataset, which is widely recognized for its use for depth estimation for driving scenarios, consists of approximately 86,000 training images. These images are captured at a rate of 10 frames per second, resulting in less than three hours of driving video. Collecting this type of driving video data is relatively straightforward and requires only a driver and a dashboard camera. In addition to continuous video data, collecting training data for monocular depth estimation can also utilize hardware, such as multi-view stereo systems or depth sensors like Time-of-Flight (ToF) and Lidar. What is common among these data sources is that they demand minimal labor and resources compared to the extensive efforts needed for ImageNet data collection. This affordability allows for the scaling up of initial model training directly within the domain of interest.

From an intuitive perspective, it is more natural to transfer knowledge between tasks that share semantic similarities, such as between different semantic tasks, than to transfer between tasks with distinct characteristics, like transitioning from a geometric task to a semantic task. The main problem is that image classification is {\em defined} by induction and therefore does not only entail, but {\em requires} a strong inductive bias, which opens the door to potentially pernicious side-effects. Induction is required because, continuing the example of the image labeled as ``cow'', there is nothing in the image of a scene that would enable one to infer the three-letter word ``cow.'' The only reason we can do so is because the present image resembles, in some way implicitly defined by the training process, {\em different images, of different scenes,} that some human has tagged with the word ``cow.'' However, since images, no matter how many, are infinitely simpler than even a single scene, say the present one, this process is not forced to learn that the extant world even exists. Rather, it learns regularities in images, each of a different scene since current models are trained with data aiming to be as independent as possible,  agnostic of the underlying scene.

Now, one may object that monocular depth estimation is itself undecidable. This is why monocular depth estimators are trained with either multiple views (motion or stereo), or with some form of supervision or partial backing from additional modalities, such as sparse depth from lidar \cite{kuznietsov2017semi} or other range sensor \cite{smisek20133d}. Then, a depth estimate is just a statistic of the learned prior. As a result, depth estimation networks should never produce {\em one} depth estimate for each image, but rather a distribution over depth maps, conditioned on the given image  \cite{yang2018conditional,yang2019dense}. Given an image, every depth map is possible, but they are not equally likely. The posterior over depth given an image then acts as a prior in depth inference using another modality, {\em e.g.,} unsupervised depth completion. The mode of this conditional prior can then be used as a depth estimate if one so desires, but with the proviso that whatever confidence one may place in that point estimate comes from inductive biases that cannot be validated 

One additional objection is that, since depth requires optimization to be inferred, and the choice of the loss function is a form of transductive bias, that is no less arbitrary than inductive bias. But this is fundamentally not the case, for the optimization residual in transductive inference refers to the data {\em here and now}, and not to data of different images in different scenes. In other words, the optimization residual is informative of the confidence of our estimate, unlike the discriminant from an inductively-trained classifier \cite{der2009aleatory}.

In some cases, one enriches (augments) the predictive loss with manually engineered transformation, calling the result ``self-supervised learning.'' Engineered transformations include small planar group transformations like translation, rotation, scaling, reflections, and range transformations such as contrast transformation or colorization. But for such group transformations, learning is not necessary since we know the general form of the maximal invariant with respect to the entire (infinite-dimensional) diffeomorphic closure of image transformations \cite{sundaramoorthi2009set}. One exception is occlusion and scale changes, which are not groups once one introduced domain quantization. So, pre-training using masking, ubiquitous in language models, does learn a preudo-invariant to occlusion, but rather than simulating occlusions, one can simply observe them in the data (a video). The only supervision is then one bit, provided temporal continuity: The fact that temporally adjacent images portray the same scene. While one would assume that video prediction as a pretraining task \cite{jin2020exploring,wu2020future,wang2020probabilistic,lao2021flow} would then yield representations with the same set of invariances to support semantics, surprisingly it does not as evident by our findings in training with videos on optical flow.

In this paper, we aim to bypass the artificial constraints imposed both by supervised classification, and self-supervision. Instead, we simply use videos to pre-train a model for depth estimation, without supervision. Then, one can use supervision to fine-tune the model for semantic segmentation. These two tasks are seemingly antipodal, yet depth estimation outperforms image classification when used as pre-training for semantic segmentation.

This also addresses one last objection that one can move to our thesis, which is that, since ImageNet data is available, it makes sense to use it. What we argue here is that, actually, it does not. The argument is corroborated by evidence: Pre-training on a geometric task improves fine-tuning a semantic task, even when compared with pre-training with a different semantic task. Nonetheless, ImageNet pre-training can be useful to expedite training depth estimation networks, likely due to the formation of generic filter banks in the early layers.

\end{document}